%% file: main.tex
\documentclass[10pt,twocolumn,letterpaper]{article}

\PassOptionsToPackage{table}{xcolor}
\usepackage{cvpr}              %
\usepackage{booktabs}
\usepackage{multirow}
\usepackage{threeparttable}
\usepackage[table]{xcolor}
\usepackage{adjustbox}

\input{preamble}
\makeatletter
  \def\title@font{\Large\bfseries}
  \let\ltx@maketitle\@maketitle
  \def\@maketitle{\bgroup%
    \let\ltx@title\@title%
    \def\@title{\resizebox{\textwidth}{!}{%
      \mbox{\title@font\ltx@title}%
    }}%
    \ltx@maketitle%
  \egroup}
\makeatother

\definecolor{cvprblue}{rgb}{0.21,0.49,0.74}
\usepackage[pagebackref,breaklinks,colorlinks,allcolors=cvprblue]{hyperref}

\def\confName{CVPR}
\def\confYear{2026}

\title{Learnability-Driven Submodular Optimization for Active Roadside 3D Detection}

\author{
Ruiyu Mao \quad Baoming Zhang \quad Nicholas Ruozzi \quad Yunhui Guo\\
The University of Texas at Dallas\\
\\
{\tt\small \{ruiyu.mao,baoming.zhang,nicholas.ruozzi,yunhui.guo\}@utdallas.edu}
}

\begin{document}
\maketitle
\input{sec/0_abstract}    
\input{sec/1_intro}
\input{sec/2_related}

\input{sec/3_preliminaries}

\input{sec/4_method}

\input{sec/5_experiments}

\input{sec/6_finalcopy}

{
    \small
    \bibliographystyle{ieeenat_fullname}
    \bibliography{main}
}

\input{sec/X_suppl}

\end{document}

%% file: sec/0_abstract.tex
\begin{abstract}
In a single roadside image, distance, blur, and occlusion can make the 3D positions of objects difficult to infer. Roadside benchmarks can use additional sensors to produce ground-truth boxes, while a monocular detector must learn from images alone. We propose LH3D, an active learning method for roadside monocular 3D object detection. It selects images in three stages based on depth confidence, class balance, and geometric variation. We evaluate LH3D on DAIR-V2X-I and Rope3D. In the reported BEVHeight comparisons, LH3D achieves higher average 3D detection AP than the uncertainty-based baselines across the Easy, Moderate, and Hard settings.
\end{abstract}

%% file: sec/1_intro.tex
\section{Introduction}
\label{sec:intro}

Modern autonomous driving systems primarily rely on ego-vehicle sensors (cameras, LiDAR, radar) to perceive their surroundings~\cite{Geiger2012CVPR,nuscenes2019,Sun_2020_CVPR}. Yet an ego-only viewpoint suffers from occlusions, intersection blind spots, and limited long-range visibility in dense traffic, motivating vehicle--infrastructure cooperation, in which infrastructure sensors provide additional viewpoints~\cite{Yu2022DAIRV2X,Ye2022Rope3D,li2022v2x}.

Among infrastructure options, roadside cameras can be deployed densely at relatively low cost and are naturally suited to bird's-eye-view (BEV) perception: camera-based detectors construct BEV representations from images using approaches that include depth-based lifting, height-based lifting, and attention~\cite{huang2021bevdet,li2022bevdepth,li2022bevformer,Yang2023BEVHeight,Wang2024BEVSpread}.

Roadside cameras are typically installed at intersections, highway ramps, and other traffic hubs, where traffic flows from multiple directions and vehicles, pedestrians, and cyclists share the same space in complex ways, leading to dense scenes with long-range targets and frequent occlusions~\cite{Ye2022Rope3D,Zhu2024RoScenes,zimmer2023tumtrafintersection,Xu2018TrafficMonitoring}. To train reliable roadside BEV detectors in such environments, large-scale datasets must capture the countless variety of objects, layouts, and conditions%
~\cite{Yu2022DAIRV2X,Ye2022Rope3D,Yang2023BEVHeight,Wang2024BEVSpread}. To label each frame, annotators must reason about 3D bounding boxes under heavy occlusion and severe depth ambiguity, cross-checking multiple visual cues and contextual references to obtain consistent labels. At city scale, the time required to annotate every frame becomes unsustainable, turning 3D labeling into the dominant cost of deploying roadside perception systems.

\begin{figure}[!t]
    \centering
    \includegraphics[width=\columnwidth]{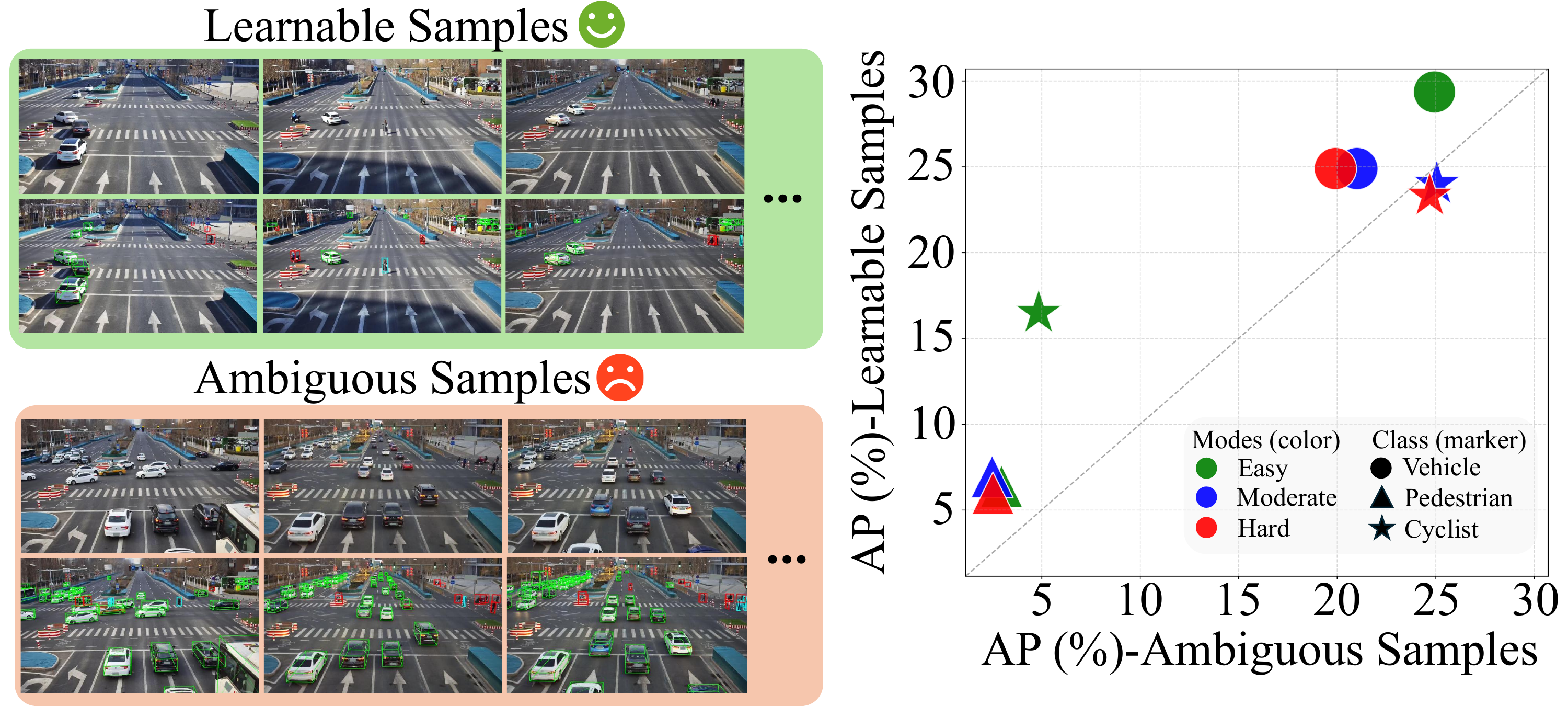}
    \caption{
    \textbf{Human study: learnable vs.\ ambiguous samples.}
    Images are categorized as \emph{learnable} or \emph{ambiguous} based on how 
    difficult they are to interpret from a single monocular view. Using this 
    partition (while training only with the dataset’s original ground-truth labels), 
    detectors trained on the ambiguous split achieve lower AP on cars and pedestrians 
    under the same annotation budget and class balance, with cyclists remaining similar. 
    This indicates that ambiguous samples provide weaker monocular supervision.
    }
    \label{fig:ambiguous_vs_easy}
\end{figure}

To reduce the cost of annotating such large data sets, we use \emph{pool-based active learning} (AL), which iteratively selects examples from an unlabeled pool for annotation~\cite{tharwat2023survey}. In our setting, these examples are roadside images. The datasets use different annotation protocols. DAIR-V2X-I contains infrastructure-side annotations. DAIR-V2X builds a separate set of cooperative annotations from paired vehicle and infrastructure frames for VIC-Sync~\cite{Yu2022DAIRV2X}. Rope3D uses point clouds from vehicle-mounted LiDAR calibrated to roadside cameras to obtain 3D labels~\cite{Ye2022Rope3D}. A detector operating on a single roadside image has fewer cues about the 3D location of distant or heavily occluded objects. These inherently ambiguous samples can be problematic in an AL setting: Traditional \emph{uncertainty-based} AL and \emph{diversity-based} AL prioritize high-uncertainty or distributionally “novel” scenes, which in roadside BEV often coincide with inherently ambiguous samples. As a result, much of the annotation budget is wasted on intrinsically unlearnable labels instead of filtering them out; Fig.~\ref{fig:ambiguous_vs_easy} shows that detectors trained on these ambiguous scenes perform worse than those trained on learnable ones under the same budget and class distribution, indicating weaker supervision.

We therefore rephrase this challenge of AL for 3D roadside BEV perception through the lens of \emph{learnability}. A single roadside view can leave the depth of an occluded object uncertain. To select from the unlabeled pool, we characterize learnability by how well a scene supports unambiguous depth estimation, balanced semantics, and informative yet resolvable geometry. Reliable depth estimation is critical for monocular BEV as it yields stable BEV geometry and accurate 3D bounding boxes, whereas ambiguous depth cascades into large localization errors and inconsistent labels. At the same time, roadside traffic is dominated by vehicles, but a safe detector cannot be biased toward the majority class; it must still learn pedestrians and cyclists robustly from limited annotations. Finally, roadside environments exhibit a wide range of layouts, e.g., different intersection shapes, lane configurations, and traffic patterns, and an effective model should be exposed to diverse but still learnable geometric configurations rather than repeatedly oversampling a few common patterns.

We decompose learnability into three complementary aspects: \textbf{(i) depth confidence}, \textbf{(ii) semantic balance}, and \textbf{(iii) geometric variation}. Depth confidence reflects how stable and predictable the image-to-depth projection is in monocular BEV lifting; samples with confident depth estimates form the foundation for learnable geometry. Semantic balance captures whether an image provides a balanced exposure across rare and common classes, preventing the model from overfitting to dominant vehicle categories and neglecting vulnerable road users. Geometric variation quantifies how the spatial configuration of objects differs across scenes in the roadside BEV scene space, encouraging exploration of novel yet still learnable layouts. Together, these criteria form a unified measure of learnability, allowing our framework to select samples that are not only informative and diverse, but also \emph{reliably learnable} and less likely to be inherently ambiguous. To jointly optimize these three learnability factors in a principled way, we formulate active selection as a concave-over-modular submodular maximization problem, which provides a natural structure for modeling balanced coverage and supports stage-wise greedy optimization on fixed candidate pools~\cite{nemhauser1978analysis,wei2014submodular,bilmes2022submodularity,mirzasoleiman2015lazier,krause2008near}.

To the best of our knowledge, this is the first work to explicitly identify inherently ambiguous samples in roadside BEV perception and to design a systematic learnability-driven active learning framework that directly addresses them. Our contributions can be summarized as follows.
\begin{itemize}[leftmargin=*,noitemsep,topsep=0pt]
    \item We identify and formalize \emph{inherently ambiguous samples} in roadside BEV perception, showing how single-view, single-modality constraints systematically lead to samples that are fundamentally unlearnable.
    \item We introduce a \emph{learnability-driven formulation} for active learning, grounded in three complementary factors—depth confidence, semantic balance, and geometric variation—that together characterize when a monocular roadside scene can be reliably learned.
    \item We propose \textbf{LH3D} (\textbf{L}earnable \textbf{H}ierarchical \textbf{3D}), a \emph{three-stage submodular active learning framework} that operationalizes these learnability factors through concave-over-modular objectives, enabling efficient greedy selection that suppresses inherently ambiguous samples while maintaining semantic and geometric coverage.
\end{itemize}

%% file: sec/2_related.tex
\section{Related Work}
\label{sec:related}
\textbf{Active Learning (AL).} Active learning selects examples from an unlabeled pool for annotation. Image-classification approaches include adversarial selection, cost-effective sampling, core-set selection, and predicted-loss ranking~\cite{sinha2019variational,wang2016cost,sener2017active,yoo2019learning}. Uncertainty sampling can use a classifier's predictions~\cite{lewis1995sequential,wang2014new}, including their Shannon entropy~\cite{lin1991divergence}. Bayesian CNNs can estimate predictive uncertainty with dropout~\cite{gal2015bayesian}, while BALD selects examples by the mutual information between their labels and the model parameters~\cite{houlsby2011bayesian}. Other selection criteria include expected variance reduction~\cite{cohn1996active} and expected error reduction on a graph~\cite{mac2014hierarchical}. Diversity-based methods seek coverage of the data distribution~\cite{agarwal2020contextual,sener2017active}. Several approaches combine uncertainty or task information with diversity~\cite{yang2015multi,ash2019deep,kim2021task}. BADGE samples separated, high-magnitude gradient embeddings~\cite{ash2019deep}.

\noindent
\textbf{AL for Roadside BEV Perception.}
Roadside BEV perception is an emerging yet under-explored direction in autonomous driving. Existing works such as BEVHeight~\cite{Yang2023BEVHeight} demonstrate that exploiting height distributions, rather than depth alone, can substantially improve BEV reasoning from fixed infrastructure cameras. However, training such models remains annotation-intensive due to the large number of small, distant objects and the geometric ambiguity inherent in monocular roadside views. Active learning (AL) has been widely studied for image classification and regression, and recent efforts for object detection explore committee-based sampling~\cite{roy2018deep}, consistency-based acquisition~\cite{yu2022consistency}, evidential hierarchical uncertainty~\cite{park2023active}, and plug-in scoring modules for large detectors~\cite{yang2024plug}. Still, AL for \emph{3D} detection is far less developed, and almost no prior work examines AL in the context of roadside BEV perception, where depth ambiguity, long-tail semantics, and geometric variability create unique difficulties not addressed by conventional AL heuristics.

%% file: sec/3_preliminaries.tex
\section{Preliminaries}
\label{sec:preliminaries}
We consider BEV 3D object detection from monocular roadside cameras.
Let $\{I_i\}_{i=1}^N$ denote the set of roadside images, where $I_i$ is acquired by a static roadside sensor with known intrinsic matrix $K$ and extrinsic matrix $E$.
For each image $I_i$, we denote by $\mathcal O_i=\{o_{ij}\}_{j=1}^{M_i}$ the set of foreground objects in the scene, and by $\mathcal C=\{c_1,c_2,\ldots,c_n\}$ the set of semantic categories in a global ground coordinate system.
The goal of the detector is to predict, for every object $o_{ij}$, its 3D bounding box $B_{ij}$ and semantic label $c_{ij}\in\mathcal C$. Each 3D bounding box is parameterized as
\begin{equation}
B_{ij} = (x_{ij}, y_{ij}, z_{ij},\, d_{ij}^x, d_{ij}^y, d_{ij}^z,\, \psi_{ij}),
\label{eq:box-param}
\end{equation}
where $(x_{ij},y_{ij},z_{ij})$ is the box center in the global ground coordinate system, $(d_{ij}^x,d_{ij}^y,d_{ij}^z)$ are the side lengths along the three axes, and $\psi_{ij}$ is the yaw angle.

\subsection{BEV Perception for Roadside 3D Detection}

We adopt the existing roadside BEV detector $f_\theta$ based on a generic \emph{lift–splat (LSS-style)} BEV pipeline~\cite{philion2020lss,huang2021bevdet,li2022bevdepth}. This framework, compatible with any LSS-based backbone, consists of four main components: an image encoder, a depth projector, a BEV transformer, and a 3D detection head.

The image encoder uses a ResNet--FPN backbone to extract high-dimensional multi-scale features
$F^{\mathrm{img}}\in\mathbb R^{C_{\mathrm{img}}\times\frac{H}{16}\times\frac{W}{16}}$
from the monocular roadside image $I_i\in\mathbb R^{3\times H\times W}$, where $C_{\mathrm{img}}$ denotes the channel number.
Given the calibration matrices $(E,K)$, the depth projector predicts a per-location probability distribution $\Pi_i(d\mid u)$ over $D$ discretized depth bins and a context feature map
$F^{\mathrm{ctx}}\in\mathbb R^{C_{\mathrm{ctx}}\times\frac{H}{16}\times\frac{W}{16}}$.
Following the LSS paradigm~\cite{philion2020lss}, the two are fused into \emph{2.5D frustum features}
$F^{2.5\mathrm{D}}\in\mathbb R^{C_{\mathrm{ctx}}\times D\times\frac{H}{16}\times\frac{W}{16}}$, representing the joint appearance--geometry embedding of each camera ray.

Using $(E,K)$ and the discretized bin geometry, the frustum features are lifted into 3D space and projected onto the ground coordinate system, yielding voxelized features
$F^{3\mathrm{D}}\in\mathbb R^{X\times Y\times Z\times C_{\mathrm{ctx}}}$.
A differentiable \emph{voxel pooling} operation along the vertical dimension aggregates them into a unified BEV feature map $F^{\mathrm{bev}}\in\mathbb R^{C_{\mathrm{ctx}}\times X\times Y}$.
Finally, a BEV detection head built on $F^{\mathrm{bev}}$ predicts the set of 3D bounding boxes and labels for image $I_i$,
$\widehat{\mathcal Y}_i=\{(B_{ij},c_{ij})\}_{j=1}^{\widehat N_i}$.
Since the intrinsic and extrinsic parameters $(K,E)$ of the roadside cameras are fixed and known after installation, the overall detection process can be written compactly as
\begin{equation}
\label{eq:detector}
\widehat{\mathcal{Y}}_i = f_\theta(I_i,\,K,\,E).
\end{equation}

\subsection{Active Learning Formalism}

Let $\mathcal U$ denote the index set of unlabeled images and $\mathcal L$ the index set of labeled images, with $I_i$ the image corresponding to index $i$.
Each unlabeled image $I_i$ with $i\in\mathcal U$ contains a set of 3D objects $\mathcal O_i=\{o_{ij}\}_{j=1}^{M_i}$ that are annotated only after the image is selected.
The active learning process proceeds for $Q$ rounds. At round $q\in\{1,\ldots,Q\}$, we are given a per-round annotation budget $k_q$ measured at the \emph{image level}, meaning we must select exactly $|S_q|=k_q$ image indices, where $S_q\subseteq\mathcal U$.
Additionally, the entire active learning process is constrained by a \emph{total object budget} $K_{\text{total}}$, which limits the cumulative number of annotated objects across all rounds:
$\sum_{q=1}^{Q}\sum_{i\in S_q}\lvert\mathcal O_i\rvert \;\le\; K_{\text{total}}.$

Selection is performed greedily: at each round we iteratively add image indices until $|S_q|=k_q$.
After each round, the labeled index set is updated as $\mathcal L\leftarrow\mathcal L\cup S_q$ and the detector $f_\theta$ is retrained from the previous checkpoint using the images $\{I_i\mid i\in\mathcal L\}$.

\subsection{Submodular Functions}
A set function $F:2^{\mathcal U}\!\rightarrow\!\mathbb{R}$ on a finite ground set $\mathcal U$ is \emph{submodular} if it satisfies diminishing returns:
\begin{equation} 
\label{eq:submodular} 
F(A\cup\{i\}) - F(A) \;\ge\; F(B\cup\{i\}) - F(B), 
\end{equation} 
for all $A\subseteq B\subseteq \mathcal U$ and $i\notin B$.
When $F$ is normalized at the empty set and also monotone, i.e., $A\subseteq B\Rightarrow F(A)\le F(B)$,
a greedy algorithm that iteratively adds the element with largest marginal gain
$\Delta(i\mid S)=F(S\cup\{i\})-F(S)$ achieves a $(1-1/e)$ approximation
under a cardinality constraint $|S|\le k$~\cite{nemhauser1978analysis}.
Such objectives naturally model \emph{coverage}, \emph{diversity}, and \emph{information gain} in data subset selection.

We repeatedly use a concave-over-modular family defined by nonnegative weights
$w_{i,\omega}\ge0$ and a non-decreasing concave function
$\phi:\mathbb{R}\rightarrow\mathbb{R}$:
\begin{equation}
\label{eq:concave-over-modular}
F_{\mathrm{cov}}(S) \;=\; \sum_{\omega\in\Omega} \phi\!\left( \sum_{i\in S} w_{i,\omega} \right),
\end{equation}
where $\Omega$ indexes coverage domains (e.g., depth, semantics, geometry).
Because $\sum_{i\in S} w_{i,\omega}$ is modular in $S$ and $\phi$ is concave,
$F_{\mathrm{cov}}$ is monotone submodular and captures \emph{balanced coverage}:
the gain from adding an element saturates once its corresponding bins are well covered.

%% file: sec/4_method.tex
\section{Method}
\label{sec:method}

\begin{figure*}[t!]
    \centering
    \includegraphics[width=0.8\textwidth]{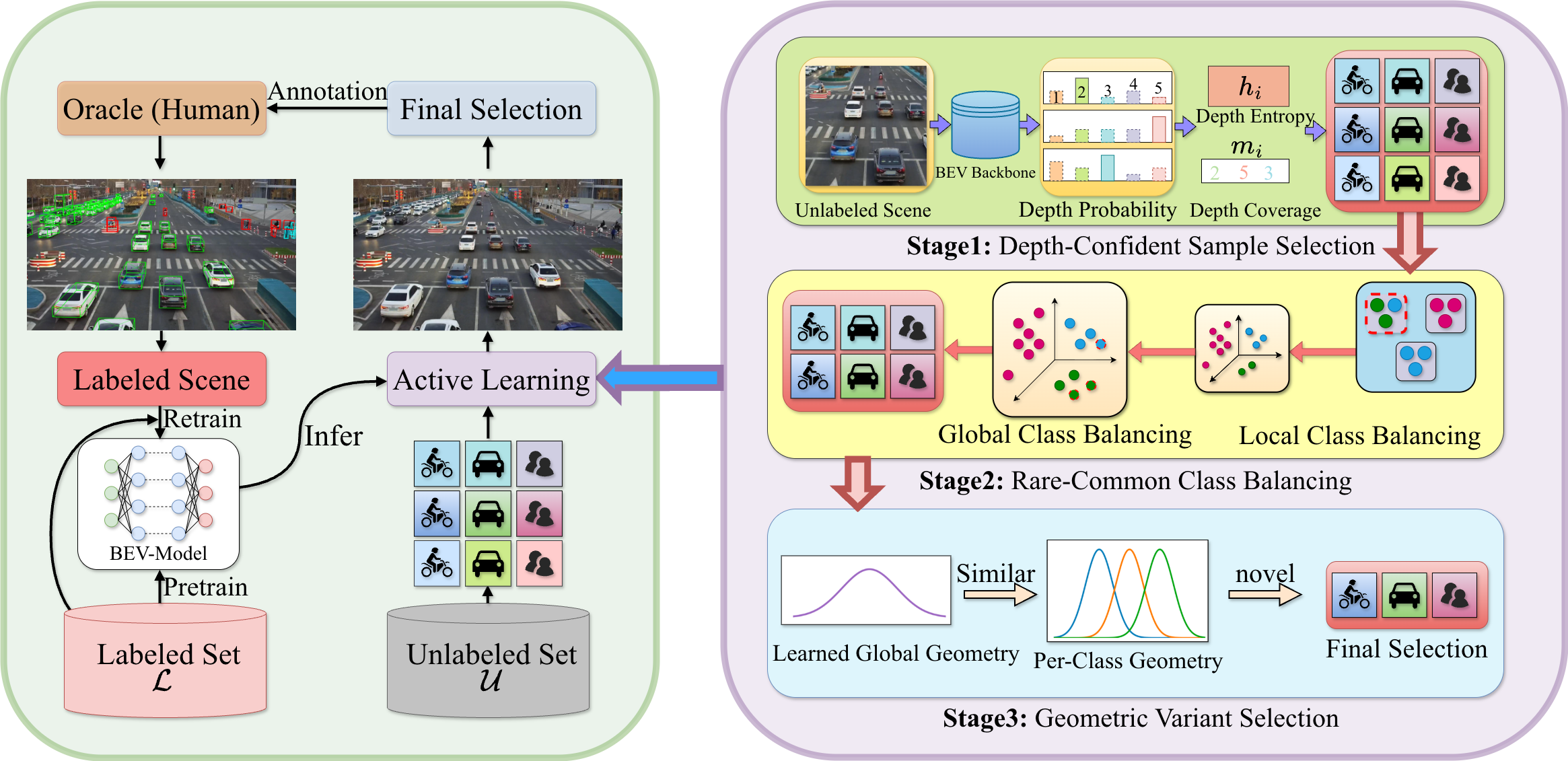}
    \caption{\textbf{Left:} Our learnability-driven active learning pipeline for roadside BEV 3D detection. 
    \textbf{Right:} The proposed LH3D three-stage selector—depth confidence, semantic balance, and geometric variation—
    which selects images that are both reliably learnable and informative for monocular roadside perception.}
    \label{fig:al_pipeline}
\end{figure*}

To tackle learnability-aware roadside BEV 3D detection, we adopt a unified selection framework grounded in the concave-over-modular formulation in Eq.~\eqref{eq:concave-over-modular}. As discussed in Sec.~\ref{sec:intro}, a roadside scene is learnable if (i) its depth can be estimated confidently, (ii) its semantic content does not aggravate class imbalance, and (iii) its per-class geometry is novel yet compatible with the learned patterns.

We encode these three aspects using monotone submodular functions
$\Phi_A$, $\Phi_B$, and $\Phi_C$:
\begin{itemize}[leftmargin=*,noitemsep,topsep=1pt]
    \item $\Phi_A$ (\textbf{Stage 1: Depth-Confident Sample Selection}) measures coverage of \emph{depth-confident} regions in the unlabeled pool, suppressing inherently ambiguous depth;
    \item $\Phi_B$ (\textbf{Stage 2: Rare–Common Class Balancing}) measures the \emph{semantic balance} of the labeled set when new images are added, discouraging dominance of majority classes (e.g., Vehicle);
    \item $\Phi_C$ (\textbf{Stage 3: Geometric Variant Selection}) measures \emph{geometric variation}, encouraging exposure to diverse but still learnable per-class spatial layouts.
\end{itemize}

At round $q$, we choose $S_q$ to improve learnability along all three axes:
\begin{equation}
\label{eq:learnability-objective}
\resizebox{\linewidth}{!}{$
\begin{aligned}
\max_{S_q \subset \mathcal U} \;
F(S_q)
&=
\underbrace{\big[\Phi_A(S_q) - \Phi_A(\mathcal U)\big]}_{\text{Depth-Confident Coverage}}
\;+\;
\underbrace{\big[\Phi_B(\mathcal L_q \cup S_q) - \Phi_B(\mathcal L_q)\big]}_{\text{Semantic Balance Gain}} \\
&\quad+\;
\underbrace{\big[\Phi_C(\mathcal L_q \cup S_q) - \Phi_C(\mathcal L_q)\big]}_{\text{Geometric Variation Gain}}.
\end{aligned}
$}
\end{equation}
The first term encourages $S_q$ to capture the depth-confident structure of $\mathcal U$, while the latter two ensure that, once $S_q$ is annotated and merged into $\mathcal L_q$, semantic balance and geometric coverage are improved rather than degraded.

Instead of maximizing Eq.~\eqref{eq:learnability-objective} over all unlabeled images, we realize $F$ as a three-stage hierarchical selector over depth confidence, semantic balance, and geometric variation. Each term is instantiated using the concave-over-modular template in Eq.~\eqref{eq:concave-over-modular}, so the active selection problem becomes a concave-over-modular submodular maximization task, which naturally models balanced coverage and admits a greedy approximation bound for each fixed candidate pool when its stage objective has nonnegative feature weights~\cite{nemhauser1978analysis,wei2014submodular,bilmes2022submodularity,mirzasoleiman2015lazier,krause2008near}.

\subsection{Stage 1: Depth-Confident Sample Selection}

Stage~1 focuses on the most basic prerequisite for monocular BEV: \emph{reliable depth}. If the depth along a camera ray is inherently ambiguous, the lifted BEV features and downstream 3D boxes are all unreliable, no matter how good the semantic or geometric scoring is. We therefore first favor images whose predicted depth distributions are confident, and among those, we prefer subsets that cover a wide range of depths.

Recall from Sec.~\ref{sec:preliminaries} that, given calibration matrices $(E,K)$, the depth projector predicts for each image $I_i$ a per-location probability distribution $\Pi_i(d\mid u)$ over $D$ discretized depth bins and a context feature map
$F^{\mathrm{ctx}}_i\in\mathbb R^{C_{\mathrm{ctx}}\times\frac{H}{16}\times\frac{W}{16}}$.
Here $u$ indexes spatial locations on $F^{\mathrm{ctx}}_i$, and $\Pi_i(d\mid u)\in[0,1]$ with
$\sum_{d=1}^D \Pi_i(d\mid u)=1$.

We measure the average depth uncertainty of $I_i$ by the (normalized) Shannon entropy of these distributions:
\begin{align}
H_i
&=\;
\mathbb{E}_{u}
\Big[-\textstyle\sum_{d=1}^{D}
\Pi_i(d\mid u)\,\log \Pi_i(d\mid u)\Big], \label{eq:sa-entropy-raw}\\[2pt]
h_i
&=\;
\frac{H_i}{\log D},
\qquad h_i\in[0,1],
\label{eq:sa-entropy}
\end{align}
where lower $h_i$ value means more confidence depth.
We map $h_i$ to a \emph{reliability weight} $r_i=r(h_i)\in(0,1]$
with $r'(h)\le 0$ (e.g., $r(h)=e^{-\tau h}$), so that depth-confident images
contribute more to the objective.

To capture which depth ranges each image occupies, we summarize the most likely depth bin of every context feature. For each location $u$, we take
\[
d^\star_{i,u} = \arg\max_{d} \Pi_i(d\mid u),
\]
and build a normalized depth histogram $m_i \in \Delta^{D-1}$ by counting these argmax bins over all locations:
\begin{equation}
\label{eq:sa-hist-avg}
m_i
\;=\;
\frac{1}{M_i}
\sum_{u}
\mathbf{e}\big(d^\star_{i,u}\big),
\end{equation}
where $M_i$ is the number of spatial locations of $F^{\mathrm{ctx}}_i$ and
$\mathbf{e}(\cdot)$ is the one-hot basis vector over the $D$ depth bins.
Intuitively, $m_i$ is the empirical distribution of the depth bins that the model is most confident about in image $I_i$.

Given a candidate subset $S$ (in Stage~1 we take $S=\mathcal U$),
its confidence-weighted depth coverage vector is
\[
Z(S)
\;=\;
\sum_{i\in S} r_i\,m_i \;\in\; \mathbb{R}^D,
\]
and we denote the $d$-th component of $Z(S)$ by $Z_d(S)$.

Instantiating the concave-over-modular template in Eq.~\eqref{eq:concave-over-modular},
we define the Stage~1 objective as
\begin{equation}
\label{eq:sa-obj}
\Phi_A(S)
\;=\;
\sum_{d=1}^{D}\log\Big(\epsilon + Z_d(S)\Big),
\end{equation}
with a small constant $\epsilon>0$.
Since $\log(\cdot)$ is non-decreasing concave and $Z(\cdot)$ is additive in $S$,
$\Phi_A$ is a monotone submodular function.
The sum of logarithms encourages a \emph{balanced} distribution of reliable depth coverage across bins, so that early annotations are spent on depth-confident scenes that collectively cover near, mid, and far ranges.
Stage~1 applies greedy selection on $\Phi_A$ to obtain the depth-confident
candidate set $S_q^{A}$ for subsequent stages.

\subsection{Stage 2: Rare-Common Class Balancing}
\begin{figure}[t]
    \centering
    \includegraphics[width=0.8\columnwidth]{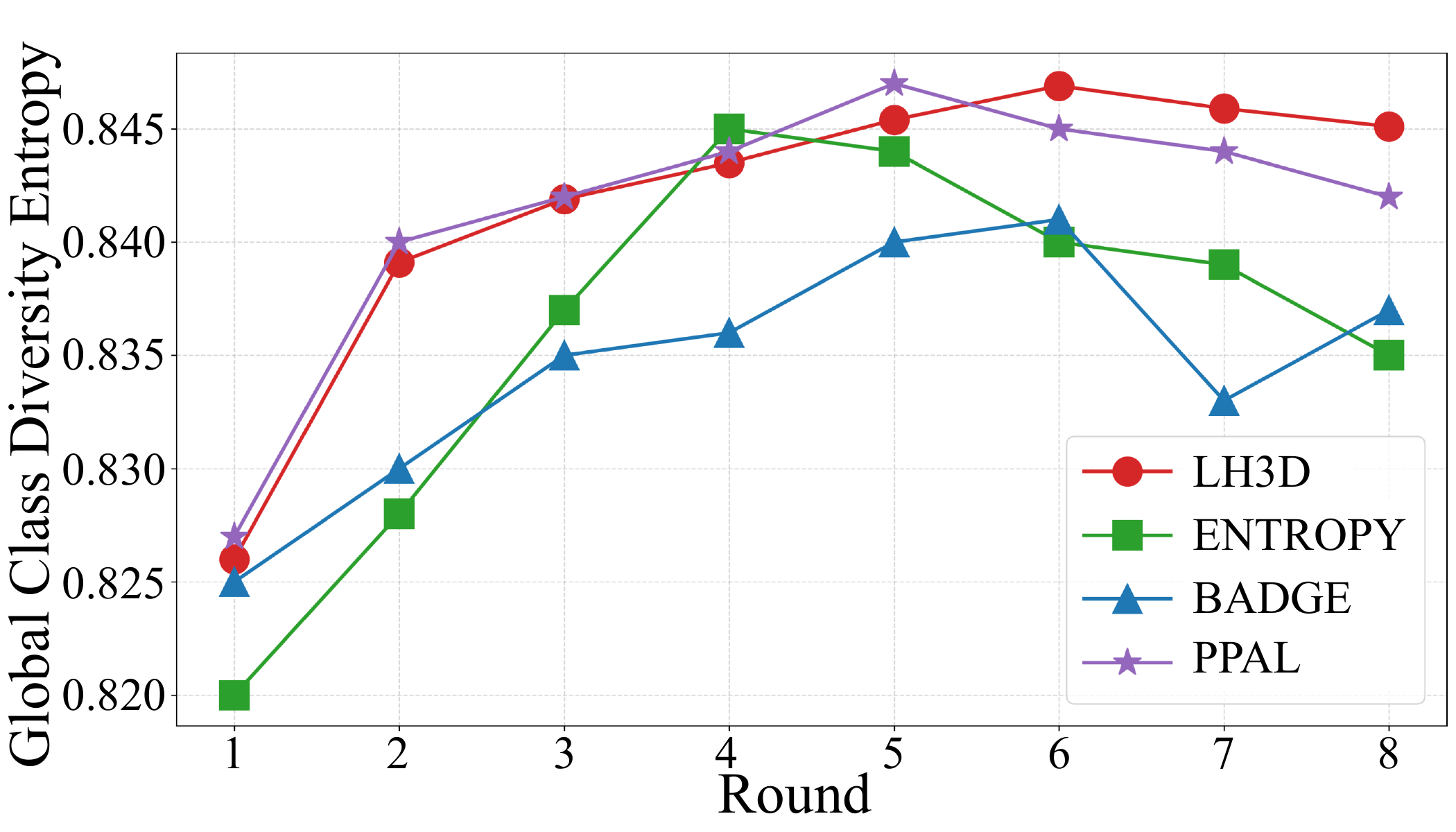}
\caption{
\textbf{Global Class Diversity Entropy across AL rounds.}
LH3D consistently achieves higher entropy than baselines, showing more balanced sampling of Car, Pedestrian, and Cyclist and preventing the increasingly imbalanced selections observed in other methods.
}
\label{fig:class-entropy}
\end{figure}

Given the depth-confident candidate set $S_q^{A}$ from Stage~1, Stage~2 targets \emph{semantic balance}. Roadside scenes are long-tailed: vehicles dominate most frames, while pedestrians and cyclists are sparse, so naively selecting “typical” traffic scenes over-trains the vehicle class and leaves rare but safety-critical categories under-represented. Stage~2 therefore favors images that (i) contain multiple classes and (ii) yield a more balanced global class distribution, as confirmed in Fig.~\ref{fig:class-entropy}.

For each image $I_i\in S_q^{A}$, we use the current detector to obtain predicted object counts $\hat{N}_{i,c}$ for each $c\in\mathcal{C}$ (optionally confidence-weighted) and normalize them into a per-image class distribution

\begin{equation}
\label{eq:stageB-entropy}
p_i(c)
=
\frac{\hat{N}_{i,c} + \beta}{\sum_{c'\in\mathcal C} (\hat{N}_{i,c'} + \beta)},
\end{equation}
where $\beta>0$ is a small smoothing constant.
We then measure the semantic diversity of $I_i$ by the entropy
\[
\delta_i
=
-\sum_{c\in\mathcal C} p_i(c)\log p_i(c),
\]
so that images with multiple classes have higher $\delta_i$.
This per-image diversity encourages us to prefer frames that already mix several classes.
We map $\delta_i$ to a nonnegative weight $\alpha_i=\alpha(\delta_i)$
(e.g., $\alpha(\delta)=1+\gamma\delta$ with $\gamma>0$), which softly emphasizes such diverse images.

To capture \emph{global} balance across the selected set, we aggregate the effective coverage of each class over a subset $S\subseteq S_q^{A}$ as
\begin{equation}
\label{eq:stageB-N}
N_c(S)
=
\sum_{i\in S} \alpha_i\,p_i(c),
\qquad c\in\mathcal{C}.
\end{equation}
Here $N_c(S)$ can be interpreted as the total exposure of class $c$
within the selected subset $S$, combining both per-image diversity ($\alpha_i$) and
how much of class $c$ each image contributes ($p_i(c)$).

We then instantiate the Stage~2 objective as a concave-over-modular function:
\begin{equation}
\label{eq:stageB-obj}
\Phi_B(S)
=
\sum_{c\in\mathcal C}
\log\big(\epsilon + N_c(S)\big),
\end{equation}
with a small $\epsilon>0$.
The logarithm induces saturation: once a class is well represented (large $N_c(S)$), its marginal gain quickly diminishes, so $\Phi_B$ favors images that contribute to underrepresented classes rather than repeatedly adding vehicle-heavy scenes. We apply greedy selection on $\Phi_B$ over $S_q^{A}$ to obtain the semantically balanced candidate set $S_q^{B}$.

\subsection{Stage 3: Geometric Variant Selection}
\label{sec:stage-c}

After Stages~1 and~2, the candidate set $S_q^{B}$ is already depth-confident and semantically balanced. Stage~3 promotes per-class \emph{geometric variation} while remaining consistent with the learned geometry.

We estimate object geometry in $\mathcal L_q$ by fitting Gaussian models to the BEV centers and heights of labeled boxes (one global and one per class), and for each unlabeled image $I_i\in S_q^{B}$ we use the average negative log-likelihood (NLL) of its predicted boxes under these Gaussians to quantify geometric novelty, discarding very low-likelihood outliers and encouraging moderate deviations. Formally, we summarize per-class BEV geometry on a coarse grid and define a geometric novelty score

\begin{equation}
s_{i,c} \;=\; -\text{NLL}\big(I_i;\mathcal N(\mu_c,\Sigma_c)\big)\;\ge0,
\end{equation}
which increases when the layout of class $c$ in $I_i$ deviates from its learned pattern
but remains within the range of the learned geometry. Aggregating these scores over a subset $S\subseteq S_q^{B}$ gives
\vspace{-0.1em} %
\begin{equation}
\label{eq:stageC-obj}
\small %
U_c(S)=\sum_{i\in S}s_{i,c}, \quad
\Phi_C(S)=\sum_{c\in\mathcal C}\log\!\big(\epsilon+U_c(S)\big),
\end{equation}
\vspace{-0.0em} %
a monotone submodular objective that favors balanced geometric coverage across classes.
Greedy maximization of $\Phi_C$ over $S_q^{B}$ yields the final annotated set $S_q$ at round $q$.

Together, Stages~1--3 form a hierarchical selector that filters out depth-ambiguous scenes, balances rare and common classes, and diversifies per-class geometry—producing a final subset $S_q$ of samples.

%% file: sec/5_experiments.tex
\section{Experiments}
\label{sec:experiments}
We briefly introduce the experiment settings and two benchmark datasets in road-side perception domain. We then compare our proposed method with state-of-the art active learning methods. Finally, We ablate our methods in detail and discuss the limitations.
\subsection{Datasets}

\textbf{DAIR-V2X}~\cite{Yu2022DAIRV2X} is a benchmark for vehicle–infrastructure cooperative driving with rich multi-modal 3D detection data. Following prior work~\cite{Yang2023BEVHeight, Wang2024BEVSpread}, we use the DAIR-V2X-I subset, which contains about 10k infrastructure-camera images and 493k 3D bounding boxes within a 0 to 200 meter range. We adopt the standard 50\%/20\%/30\% split for training, validation, and testing, and report results on the validation set since the official test annotations are unavailable.

\noindent
\textbf{Rope3D}~\cite{Ye2022Rope3D} is another large-scale benchmark for roadside 3D object detection, containing 50k images and over 1.5M annotated objects captured under diverse lighting and weather conditions across 26 intersections, with object distances up to 200 m. Following the homologous split described by Rope3D, we use 70\% of the images from each scene for training and hold out 30\% for validation. 

For evaluation, we adopt the $AP_{3D|R40}$ metric with Easy, Moderate, and Hard settings defined by box characteristics, following the KITTI~\cite{Geiger2012CVPR,simonelli2019disentangling} protocol.

\subsection{Baselines}
We compare our approach with a variety of representative active learning baselines.
1) RANDOM: A naive strategy that randomly selects samples at each round.
2) ENTROPY \cite{wang2014new}: Chooses samples with the highest predictive uncertainty measured by the entropy of posterior probabilities.
3) CORESET~\cite{sener2017active}: Employs a greedy furthest-first strategy to maximize coverage between labeled and unlabeled embeddings.
4) BADGE~\cite{ash2019deep}: Selects a batch of samples that are both informative and diverse based on gradient magnitudes.
5) BGADL~\cite{tran2019bayesian}: The original method combines active selection with model-generated training examples.

We also compare against recent AL approaches designed for 2D/3D object detection.
6) PPAL~\cite{yang2024plug}: A detector-agnostic framework that efficiently selects informative and diverse samples by combining difficulty-calibrated uncertainty with category-conditioned matching similarity.
7) HUA~\cite{park2023active}: Utilizes evidential deep learning to estimate and hierarchically aggregate uncertainties into image-level scores for more reliable sample selection.

\subsection{Implementation Details}
We use a pool-based AL setup mainly on BEVHeight~\cite{Yang2023BEVHeight}. 
At each round, the model continues training from the previous checkpoint, scores the pool, 
and we query the top 100 images. 
The labeled set starts with 500 images, and all methods follow the same total annotation 
budget of 32{,}000 objects. 
Each round is trained for 5 epochs with AdamW (lr $2\times10^{-4}$), batch size 8, on 4 RTX A5000 GPUs.

\begin{table*}[t!]
\vspace{-10pt}
\tiny
\centering
\caption{\textbf{\boldmath{$AP_{3D|R40}$} results on the DAIR-V2X-I validation set} with 20\% queried boxes. Backbones include \textit{BEVHeight}, \textit{BEVSpread}, and \textit{BEVDet}.}
\label{tab:dair-al-all}
\setlength{\tabcolsep}{6pt}
\scriptsize
\begin{threeparttable}
\resizebox{\textwidth}{!}
{
\renewcommand{\arraystretch}{0.9}
\begin{tabular}{l l ccc ccc ccc ccc}
\toprule
& & \multicolumn{3}{c}{\textbf{Vehicle (IoU=0.5)}} &
    \multicolumn{3}{c}{\textbf{Pedestrian (IoU=0.25)}} &
    \multicolumn{3}{c}{\textbf{Cyclist (IoU=0.25)}} &
    \multicolumn{3}{c}{\textbf{Average}} \\
\cmidrule(lr){3-5}\cmidrule(lr){6-8}\cmidrule(lr){9-11}\cmidrule(lr){12-14}
\textbf{Backbone} & \textbf{Method} &
Easy & Mod. & Hard &
Easy & Mod. & Hard &
Easy & Mod. & Hard &
Easy & Mod. & Hard \\
\midrule

\multirow{9}{*}{\textbf{BEVHeight}}
& RANDOM      & 61.90 & 51.37 & 51.41 & 13.63 & 13.23 & 13.42 & 30.04 & 38.70 & 39.38 & 35.19 & 34.43 & 34.74\\
& ENTROPY     & 63.42 & 54.42 & 54.51 & 17.50 & 16.57 & 16.72 & 31.45 & 36.86 & 38.57 & 37.46 & 36.67 & 36.53\\
& UNCERTAINTY & 51.77 & 44.00 & 42.52 & 13.28 & 12.60 & 12.70 & 25.72 & 30.98 & 31.56 & 30.26 & 29.86 & 28.93\\
& BGADL~\cite{tran2019bayesian}   & 63.91 & 54.77 & 54.91 & 14.97 & 14.20 & 14.19 & 27.39 & 34.07 & 35.77 & 35.42 & 34.35 & 34.96\\
& CORESET~\cite{sener2017active}  & 51.43 & 43.78 & 42.30 & 13.86 & 13.05 & 13.19 & 30.12 & 34.44 & 35.01 & 31.80 & 30.42 & 30.17\\
& BADGE~\cite{ash2019deep}        & 60.08 & 51.19 & 51.33 & 15.70 & 14.88 & 14.98 & 27.10 & 34.77 & 35.35 & 34.29 & 33.61 & 33.89\\
& PPAL~\cite{yang2024plug}        & 60.20 & 51.38 & 51.44 & \textbf{19.09} & \textbf{18.47} & \textbf{18.07} & \textbf{34.41} & 39.13 & 39.71 & 37.90 & 36.33 & 36.41\\
& HUA~\cite{park2023active}       & 60.18 & 51.37 & 51.48 & 13.98 & 13.23 & 13.33 & 30.65 & 33.84 & 34.48 & 34.94 & 32.81 & 33.10\\
\rowcolor{gray!10}
& \textbf{LH3D (Ours)}
                               & \textbf{65.36} & \textbf{56.00} & \textbf{56.03}
                               & 18.51 & 17.50 & 17.67
                               & 32.44 & \textbf{41.49} & \textbf{41.79}
                               & \textbf{38.77} & \textbf{38.33} & \textbf{38.50} \\
\midrule

\multirow{8}{*}{\textbf{BEVSpread}}
& RANDOM      & 54.00 & 54.55 & 47.51 & 14.21 & 13.96 & 13.09 & 21.20 & 32.70 & 32.81 & 29.80 & 33.74 & 31.14\\
& ENTROPY     & 59.37 & 50.66 & 50.80 & 14.35 & 13.54 & 13.67 & 24.37 & 33.10 & 33.56 & 32.70 & 32.43 & 32.68\\
& BGADL~\cite{tran2019bayesian}   & 54.14 & 48.43 & 48.44 & 15.74 & 15.05 & 14.22 & 24.89 & 32.09 & 32.72 & 31.59 & 31.86 & 31.79\\
& CORESET~\cite{sener2017active}  & 57.96 & 49.47 & 49.50 & 14.09 & 13.38 & 13.57 & 24.41 & 35.71 & 36.38 & 32.15 & 32.85 & 33.15\\
& BADGE~\cite{ash2019deep}        & 57.54 & 48.92 & 47.51 & 13.38 & 13.04 & 13.27 & 27.68 & 35.74 & 36.16 & 32.87 & 32.57 & 32.31\\
& PPAL~\cite{yang2024plug}        & 62.80 & 50.18 & 50.29 & 15.69 & 15.85 & 15.09 & 31.46 & 35.87 & 35.39 & 36.65 & 33.97 & 33.59\\
& HUA~\cite{park2023active}       & 58.97 & 49.44 & 49.54 & 16.01 & 15.75 & 15.82 & 29.87 & 30.30 & 30.77 & 34.95 & 31.83 & 32.04\\
\rowcolor{gray!10}
& \textbf{LH3D (Ours)}
                               & \textbf{63.16} & \textbf{52.45} & \textbf{52.53}
                               & \textbf{17.63} & \textbf{17.17} & \textbf{17.40}
                               & \textbf{31.77} & \textbf{37.59} & \textbf{38.28}
                               & \textbf{37.52} & \textbf{35.74} & \textbf{36.07}\\
\midrule

\multirow{8}{*}{\textbf{BEVDet}}
& RANDOM      & 56.89 & 48.46 & 48.53 & 14.68 & 14.13 & 14.12 & 21.73 & 29.73 & 29.02 & 31.00 & 31.41 & 31.56\\
& ENTROPY     & 57.55 & 48.41 & 48.40 & 15.83 & 13.82 & 12.98 & 21.97 & 32.76 & 31.75 & 31.78 & 31.66 & 31.04\\
& BGADL~\cite{tran2019bayesian}   & 55.23 & 47.68 & 47.63 & 14.75 & 14.04 & 14.16 & \textbf{23.23} & 29.61 & 29.56 & 31.07 & 30.44 & 30.45\\
& CORESET~\cite{sener2017active}  & 54.26 & 46.65 & 46.61 & 14.87 & 14.53 & 14.59 & 21.08 & 26.03 & 26.04 & 30.07 & 29.07 & 29.08\\
& BADGE~\cite{ash2019deep}        & 56.64 & 49.17 & 49.23 & 14.47 & 13.82 & 13.95 & 20.87 & 30.40 & 29.63 & 30.66 & 31.13 & 30.94\\
& PPAL~\cite{yang2024plug}        & 56.99 & \textbf{49.61} & \textbf{49.62} & 15.57 & 14.78 & 14.23 & 22.99 & 33.37 & 33.98 & 31.85 & 32.59 & 32.61\\
& HUA~\cite{park2023active}       & 57.95 & 48.84 & 48.37 & 15.12 & 14.64 & 14.66 & 21.46 & 31.46 & 31.80 & 31.51 & 31.65 & 31.61\\
\rowcolor{gray!10}
& \textbf{LH3D (Ours)}
                               & \textbf{58.98} & 48.67 & 48.77
                               & \textbf{15.83} & \textbf{14.97} & \textbf{15.06}
                               & 23.09 & \textbf{34.63} & \textbf{35.20}
                               & \textbf{32.63} & \textbf{32.76} & \textbf{33.01}\\
\bottomrule
\end{tabular}
}
\end{threeparttable}

\end{table*}

\begin{figure}[t!]
    \centering
    \includegraphics[width=\columnwidth]{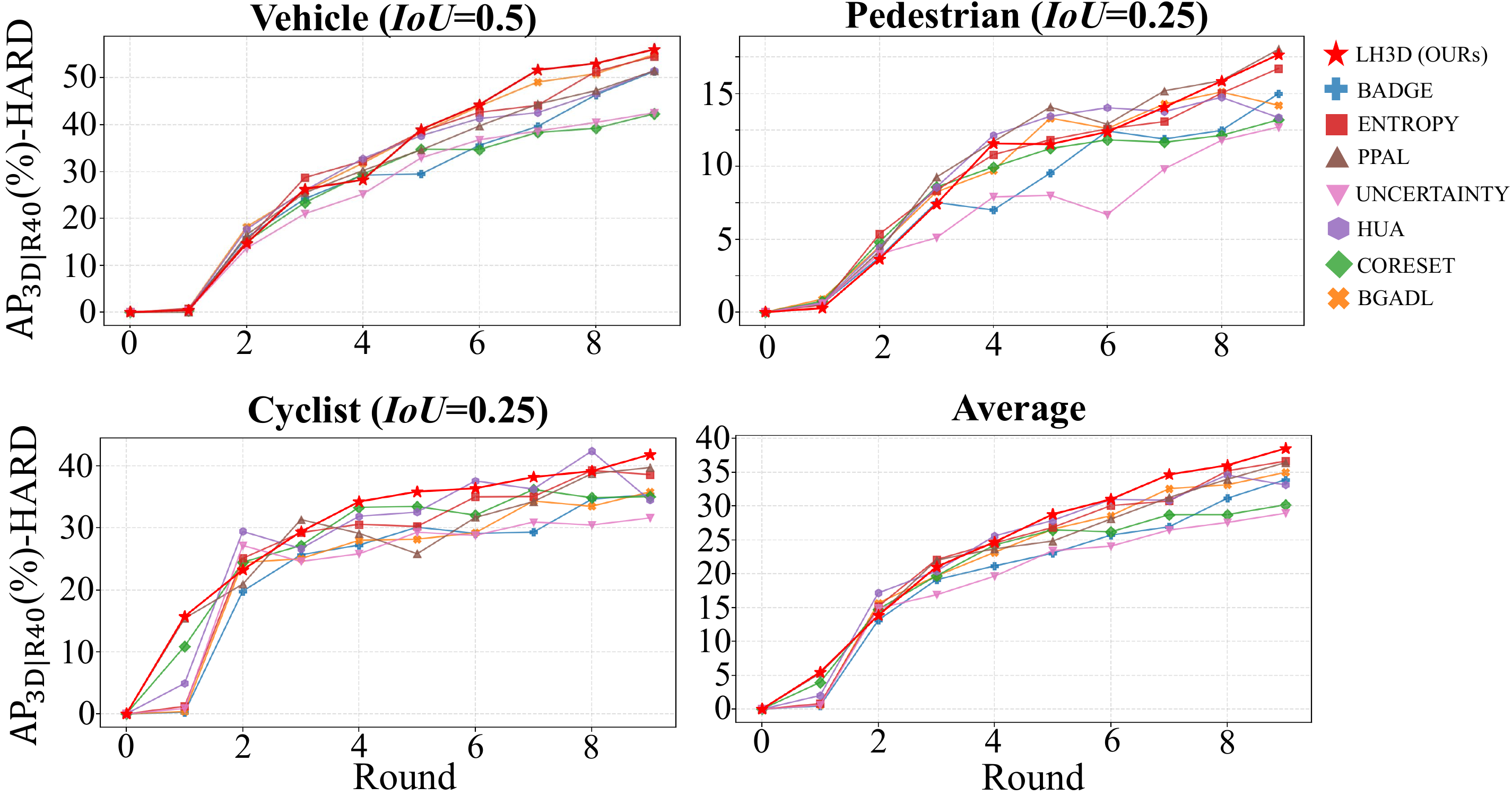}
    \caption{Active learning performance progression on the DAIR-V2X-I validation set using the BEVHeight backbone under the hard modes. 
    }
    \label{fig:round-ap}
\end{figure}

\begin{figure*}[t!]
    \centering
    \includegraphics[width=\textwidth]{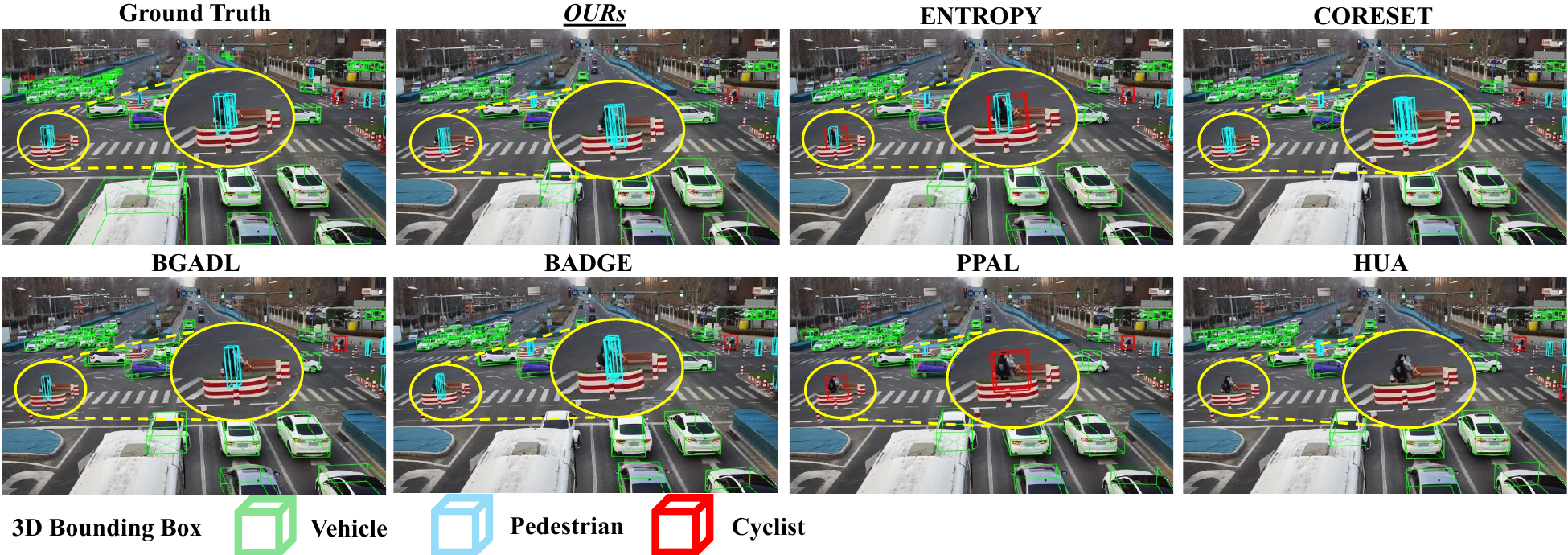}
    \caption{\textbf{Visulization results of baselines and our proposed method.} Our method (LH3D) successfully detects the pedestrian that other active learning baselines fail to identify in complex traffic environments. The 3D bounding boxes for vehicles, pedestrians, and cyclists are shown in green, blue, and red, respectively.
    }
    \label{fig:visulization}
\end{figure*}

\subsection{Results}
\textbf{DAIR-V2X Dataset Results.} We evaluate the performance of the LH3D method in Table~\ref{tab:dair-al-all}, using three different backbone detectors. LH3D achieves the highest average performance, consistently outperforming all the baseline AL methods. In particular, with BEVHeight as the backbone, LH3D achieves average 3D AP improvements of 0.87\%, 2.00\%, and 2.19\% over PPAL~\cite{yang2024plug} under the \textit{easy}, \textit{moderate}, and \textit{hard} evaluation modes. We observe improvements of 0.87\%, 5.52\%, and 5.40\% over HUA~\cite{park2023active}, respectively.

Figure~\ref{fig:round-ap} shows the performance of the AL methods across object categories and learning rounds. For the Vehicle class, our method achieves faster performance gains in the early rounds and maintains the highest 3D AP at later stages. A similar trend can be observed for Pedestrian and Cyclist, where our approach yields higher accuracy and smoother convergence.

Figure~\ref{fig:visulization} compares the 3D detection visualizations of baseline methods and our proposed LH3D framework. Our method accurately identifies pedestrians that are either missed or misclassified by competing approaches under dense traffic conditions.

\begin{table*}[t]
\centering
\vspace{-5pt}
\caption{\textbf{\boldmath{$AP_{3D|R40}$} results on the Rope3D validation set}, comparing different methods with 20\% queried boxes. \textit{\textbf{BEVHeight}} is used as the backbone detector.}
\label{tab:rope3d-al-bevheight}
\setlength{\tabcolsep}{4.5pt}
{\tiny        %
\begin{threeparttable}
\resizebox{\textwidth}{!}{
\renewcommand{\arraystretch}{0.9}
\begin{tabular}{lcccccccccccc}
\toprule
& \multicolumn{3}{c}{\textbf{Vehicle (\textit{IoU=0.5})}} &
  \multicolumn{3}{c}{\textbf{Pedestrian (\textit{IoU=0.25})}} &
  \multicolumn{3}{c}{\textbf{Cyclist (\textit{IoU=0.25})}} &
  \multicolumn{3}{c}{\textbf{Average}} \\
\cmidrule(lr){2-4}\cmidrule(lr){5-7}\cmidrule(lr){8-10}\cmidrule(lr){11-13}
\textbf{Method} & Easy & Mod. & Hard & Easy & Mod. & Hard & Easy & Mod. & Hard & Easy & Mod. & Hard \\
\midrule
RANDOM    & 22.77 & 19.70 & 19.65 & 1.52 & 1.67 & 1.50 & 12.28 & 14.81 & 14.80 & 12.19 & 12.06 & 11.99 \\
ENTROPY   & 27.25 & 23.24 & 23.16 & 1.04 & 1.00 & 1.02 & 9.18 & 12.90 & 12.96 & 12.49 & 12.38 & 12.38\\
BGADL~\cite{tran2019bayesian} & 26.50 & 22.80 & 23.10 & 1.20 & 1.05 & 1.10 & 10.00 & 12.95 & 12.50 & 12.57 & 12.27 & 12.23\\
CORESET~\cite{sener2017active} & 26.01 & 22.31 & 22.26 & 1.48 & 1.52 & 1.56 & 13.24 & 14.76 & 14.74 & 13.58 & 12.86 & 12.85\\
BADGE~\cite{ash2019deep} & 27.96 & 24.41 & 23.00 & \textbf{2.06} & \textbf{2.09} & \textbf{2.12} & 12.49 & 15.47 & 14.42 & 14.17 & 13.99 & 13.18\\
\midrule
PPAL~\cite{yang2024plug} & \textbf{29.84} & 25.55 & 24.12 & 1.95 & 1.72 & 1.73 & 12.41 & 14.89 & 14.80 & 14.73 & 14.05 & 13.55\\
HUA~\cite{park2023active} & 25.08 & 22.06 & 22.02 & 1.56 & 1.58 & 1.62 & 10.78 & 13.66 & 13.68 & 12.47 & 12.43 & 12.44\\
\midrule
\rowcolor{gray!10}
LH3D (\textbf{Ours}) & 29.77 & \textbf{27.60} & \textbf{26.12} & 1.74 & 2.00 & 2.04 & \textbf{13.39} & \textbf{17.70} & \textbf{16.69} & \textbf{14.97} & \textbf{15.77} & \textbf{14.95}\\
\bottomrule
\end{tabular}
}
\end{threeparttable}
}
\end{table*}

\noindent
\textbf{Rope3D Dataset Results.} To further test the generality of our method, we evaluate the proposed method on the Rope3D validation set, as shown in Table~\ref{tab:rope3d-al-bevheight}. Consistent with the results on the DAIR-V2X-I dataset, our approach achieves competitive or superior performance compared with existing active learning baselines across most object categories and difficulty levels. Using \textit{BEVHeight} as the backbone, LH3D achieves the highest average 3D AP among all competing methods. Compared to PPAL~\cite{yang2024plug}, our method yields improvements of 0.24\%, 2.05\%, and 1.40\% under the \textit{easy}, \textit{moderate}, and \textit{hard} settings, respectively. Furthermore, relative to HUA~\cite{park2023active}, LH3D achieves additional improvement of 2.50\%, 3.34\%, and 2.51\%.

\subsection{Ablation Study and Analysis}
\textbf{Human study on inherently ambiguous samples.}
As shown in Fig.~\ref{fig:ambiguous_vs_easy}, models trained on the ambiguous split
consistently achieve lower AP for vehicles and pedestrians across the easy, moderate, 
and hard settings, while cyclists are comparable on the moderate and hard levels and 
only slightly worse on the easy level. This indicates that ambiguous images provide 
weaker training supervision than learnable ones, especially for safety-critical 
vehicle and pedestrian categories. Additional experimental details are given in the 
supplementary material.

\begin{table}[t]
\small
\centering
\caption{Ablation study on stage ordering of our LH3D framework using the BEVHeight backbone (Hard setting). 
DC = Depth Confidence, SB = Semantic Balance, GV = Geometric Variation.}
\vspace{-5pt}
\resizebox{\columnwidth}{!}{
\begin{tabular}{ccccc}
\toprule
Order & \textbf{Car} & \textbf{Pedestrian} & \textbf{Cyclist} & \textbf{Average}\\ 
\midrule
DC–GV–SB & 50.62 & 16.83 & 37.10 & 34.85\\
SB–GV–DC & 51.82 & 15.14 & 36.81 & 34.59\\
SB–DC–GV & 55.90 & 12.46 & 35.95 & 34.77\\
GV–DC–SB & 40.04 & 13.02 & 32.67 & 28.58\\
GV–SB–DC & 47.31 & 15.16 & 36.63 & 33.03\\

\rowcolor{gray!10}
\textbf{Ours (DC–SB–GV)} & \textbf{56.03} & \textbf{17.67} & \textbf{41.79} & \textbf{38.50} \\ 
\bottomrule
\end{tabular}}
\vspace{-0.3cm}
\label{tab:stage_order}
\end{table}

\noindent
\textbf{Ablation on stage ordering.}
LH3D places depth confidence (DC) first, semantic balance (SB) second, and geometric variation (GV) last. As shown in Table~\ref{tab:stage_order}, our DC–SB–GV ordering outperforms all $3!$ permutations, while moving DC out of the first stage or starting with GV consistently degrades performance, confirming the necessity of this priority design.

\noindent
\textbf{Ablation on semantic balance across AL rounds.}
Figure~\ref{fig:class-entropy} reports the class-entropy of all accumulated selections at each AL round. LH3D maintains consistently higher entropy than ENTROPY, BADGE, and PPAL, indicating more balanced class exposure throughout selection and validating the effectiveness of our Stage~2 design.

\noindent
\textbf{Ablation on annotation budget.}
We also explored different annotation budgets and observed that performance 
saturates quickly: with our 32K-object budget, LH3D already reaches about 80\% of the 
fully supervised performance. We therefore report results under this budget in the main 
paper (for full budget curves, see the supplementary material).

%% file: sec/6_finalcopy.tex
\section{Conclusion}
We show that inherently ambiguous samples form a key bottleneck for active learning in roadside monocular 3D detection. LH3D addresses this by selecting data via depth confidence, semantic balance, and geometric variation within a unified submodular objective. On DAIR-V2X-I~\cite{Yu2022DAIRV2X} and Rope3D~\cite{Ye2022Rope3D}, this learnability-based selection improves 3D detection under reduced annotation budgets, indicating that learnability is a more practical target than raw uncertainty for roadside BEV perception.

%% file: sec/X_suppl.tex
\clearpage
\maketitlesupplementary

\section{Supplementary Material}
\label{sec:supplementary}

Due to space limitations in the main manuscript, we provide additional theoretical proofs, detailed dataset specifications, and extensive experimental analyses in this supplementary material. The content is organized as follows:

\begin{itemize}
    \item \textbf{Theoretical Analysis (Sec.~\ref{sec:theory}):} We provide the formal mathematical proofs regarding the monotonicity and submodularity of our proposed objective functions, guaranteeing the theoretical efficiency of the greedy optimization used in LH3D.
    
    \item \textbf{Dataset Details (Sec.~\ref{sec:dataset}):} We provide detailed specifications for the primary evaluation dataset, \textbf{DAIR-V2X-I}, and the generalization dataset, \textbf{Rope3D}. We clarify the evaluation protocol, which employs the standard \textbf{$AP_{3D|R40}$ metric} across KITTI-style difficulty levels.
    
    \item \textbf{Failure Case Analysis (Sec.~\ref{sec:failure}):} We analyze typical failure modes, highlighting issues with long-range vehicles and occluded pedestrians/cyclists (fragmentation and misclassification). This analysis underscores the inherent limitations of monocular 3D estimation under extreme distance and visual ambiguity.
    
    \item \textbf{Validation of Hierarchical Stages (Sec.~\ref{sec:stage_validity}):} We present in-depth discussions and empirical evidence (including visualization and metric analysis) to demonstrate the necessity and effectiveness of each individual stage in our three-stage learnability framework.
    
    \item \textbf{Ablation Studies on Annotation Budgets (Sec.~\ref{sec:budget_ablation}):} We report extended performance comparisons across a wider range of annotation budgets (from low-budget to high-budget regimes) to verify the robustness of LH3D.

    \item \textbf{Generalization Experiments (Sec.~\ref{sec:rope3D}):} We extend our evaluation to the Rope3D dataset and test across different detector architectures (BEVSpread and BEVDet) to demonstrate the generalization ability of our method beyond a specific setup.
    
    \item \textbf{Computational Complexity (Sec.~\ref{sec:complexity}):} We analyze the time complexity of our selection algorithm, showing that the computational overhead is negligible compared to the training costs.

    \item \textbf{Extended Analysis: Human Study (Sec.~\ref{sec:human_study}):} We detail the controlled human study that isolates the impact of inherent ambiguity, empirically proving that ambiguous samples provide weaker supervision than learnable ones even with perfect ground truth.
\end{itemize}

\subsection{Theoretical Analysis}
\label{sec:theory}

In this section, we provide the formal proof that the objective functions proposed in our LH3D framework---specifically $\Phi_A$ (Depth Confidence), $\Phi_B$ (Semantic Balance), and $\Phi_C$ (Geometric Variation)---are monotone submodular. For each fixed candidate pool and cardinality budget, exact greedy maximization of a monotone submodular stage objective achieves at least $(1-1/e)$ of the optimal gain above its empty-set value~\cite{nemhauser1978analysis}. The bound applies separately within each stage.

\subsection{Definitions}

Let $\mathcal{U}$ be the finite ground set of unlabeled images. A set function $F: 2^{\mathcal{U}} \to \mathbb{R}$ maps a subset $S \subseteq \mathcal{U}$ to a real value.

\vspace{5pt}
\noindent\textbf{Definition 1 (Monotonicity).} A set function $F$ is monotone if for all subsets $A \subseteq B \subseteq \mathcal{U}$, it holds that $F(A) \le F(B)$.

\vspace{5pt}
\noindent\textbf{Definition 2 (Submodularity).} A set function $F$ is submodular if it satisfies the property of diminishing returns. Formally, for all $A \subseteq B \subseteq \mathcal{U}$ and any element $x \in \mathcal{U} \setminus B$:
\begin{equation}
    F(A \cup \{x\}) - F(A) \ge F(B \cup \{x\}) - F(B).
\end{equation}

\subsection{Submodularity of Concave-Over-Modular Functions}

Our learnability objectives are formulated using the \textit{concave-over-modular} template defined in Eq.~\eqref{eq:concave-over-modular} of the main paper. We now prove that functions of this form are monotone submodular.

\vspace{5pt}
\noindent\textbf{Theorem 1.} 
Let $w_i \ge 0$ be a non-negative weight associated with each element $i \in \mathcal{U}$. Let $g(S) = \sum_{i \in S} w_i$ be a modular function, and let $\phi: \mathbb{R}_{\ge 0} \to \mathbb{R}$ be a non-decreasing, concave function. Then, the composite function $F(S) = \phi(g(S))$ is monotone submodular.

\vspace{5pt}
\noindent\textit{Proof.}
\textbf{Monotonicity:} Since $w_i \ge 0$, if $A \subseteq B$, then $g(A) \le g(B)$. Because $\phi$ is non-decreasing, it follows that $\phi(g(A)) \le \phi(g(B))$. Thus, $F(S)$ is monotone.

\textbf{Submodularity:} Let $A \subseteq B \subseteq \mathcal{U}$ and $x \in \mathcal{U} \setminus B$. Let $\Delta = w_x \ge 0$ be the weight of the new element. We define the values of the modular function as $v_A = g(A)$ and $v_B = g(B)$. Since $A \subseteq B$ and weights are non-negative, we have $v_A \le v_B$.
The marginal gain of adding $x$ to $A$ is:
\begin{equation}
    \Delta F(x \mid A) = \phi(v_A + \Delta) - \phi(v_A).
\end{equation}
Similarly, the marginal gain for $B$ is:
\begin{equation}
    \Delta F(x \mid B) = \phi(v_B + \Delta) - \phi(v_B).
\end{equation}
Since $\phi$ is a concave function, its gradients (or discrete increments) are non-increasing. Therefore, for $v_A \le v_B$ and any increment $\Delta \ge 0$, the inequality
\begin{equation}
    \phi(v_A + \Delta) - \phi(v_A) \ge \phi(v_B + \Delta) - \phi(v_B)
\end{equation}
holds. This satisfies the definition of submodularity. \hfill

\subsection{Application to LH3D Objectives}

We apply Theorem 1 to the three stages of our framework.

\textbf{Closure under Summation.} We first note that a non-negative linear combination of submodular functions is also submodular. That is, if $F_1, \dots, F_k$ are submodular, then $F(S) = \sum_k F_k(S)$ is submodular.

\begin{itemize}
    \item \textbf{Stage 1: Depth-Confident Sample Selection (Eq.~\ref{eq:sa-obj}):} 
    $\Phi_A(S) = \sum_{d=1}^{D} \log(\epsilon + Z_d(S))$. 
    Here, $Z_d(S) = \sum_{i \in S} r_i m_{i,d}$ is a modular sum with non-negative weights $r_i m_{i,d}$. The function $\phi(z) = \log(\epsilon + z)$ is concave and non-decreasing for $z \ge 0$ (given $\epsilon > 0$). Thus, each term is submodular, and their sum $\Phi_A$ is submodular.
    
    \item \textbf{Stage 2: Rare-Common Class Balancing (Eq.~\ref{eq:stageB-obj}):} 
    $\Phi_B(S) = \sum_{c \in \mathcal{C}} \log(\epsilon + N_c(S))$.
    Similarly, $N_c(S) = \sum_{i \in S} \alpha_i p_i(c)$ is a modular coverage term. By the same logic, $\Phi_B$ is a sum of concave-over-modular functions and is therefore submodular.
    
    \item \textbf{Stage 3: Geometric Variant Selection (Eq.~\ref{eq:stageC-obj}):}
    $\Phi_C(S) = \sum_{c \in \mathcal{C}} \log(\epsilon + U_c(S))$.
    With $U_c(S) = \sum_{i \in S} s_{i,c}$ being modular (sum of novelty scores), $\Phi_C$ is also submodular.
\end{itemize}

\noindent \textbf{Conclusion:} For fixed nonnegative feature weights, each stage objective is monotone submodular. Under a cardinality constraint, the approximation bound applies to each normalized stage objective on its own candidate pool.

\subsection{Datasets}
\label{sec:dataset}
\textbf{DAIR-V2X}~\cite{Yu2022DAIRV2X} is a large-scale benchmark for vehicle–infrastructure cooperative autonomous driving, offering a rich multi-modal 3D object detection resource. Following prior work~\cite{Yang2023BEVHeight, Wang2024BEVSpread}, we focus on the DAIR-V2X-I subset, which comprises approximately 10k images captured from infrastructure-mounted cameras to study roadside perception. The subset includes 493k 3D bounding box annotations spanning distances from 0 to 200 meter. We adopt the standard data split of 50\%, 20\%, and 30\% for training, validation, and testing, respectively. As the official test annotations are not yet released, all evaluations are conducted on the validation set.

\noindent
\textbf{Rope3D}~\cite{Ye2022Rope3D} is another benchmark for roadside 3D object detection. It comprises 50 k images and over 1.5 M 3D object annotations captured under diverse conditions, including varying lighting (day, night, dusk) and weather (rainy, sunny, cloudy) across 26 distinct intersections, with object distances ranging from 0 m to 200 m. Following the homologous split described by Rope3D, we use 70\% of the images from each scene for training and hold out 30\% for validation. 

For validation metrics, we leverage $AP_{3D|R40}$ metric to evaluate 3D bounding boxes. The results are reported in three difficulty levels—Easy, Moderate, and Hard—based on box characteristics, following the KITTI ~\cite{Geiger2012CVPR,simonelli2019disentangling} evaluation protocol.

\begin{figure*}[t!]
    \centering \includegraphics[width=0.8\textwidth]{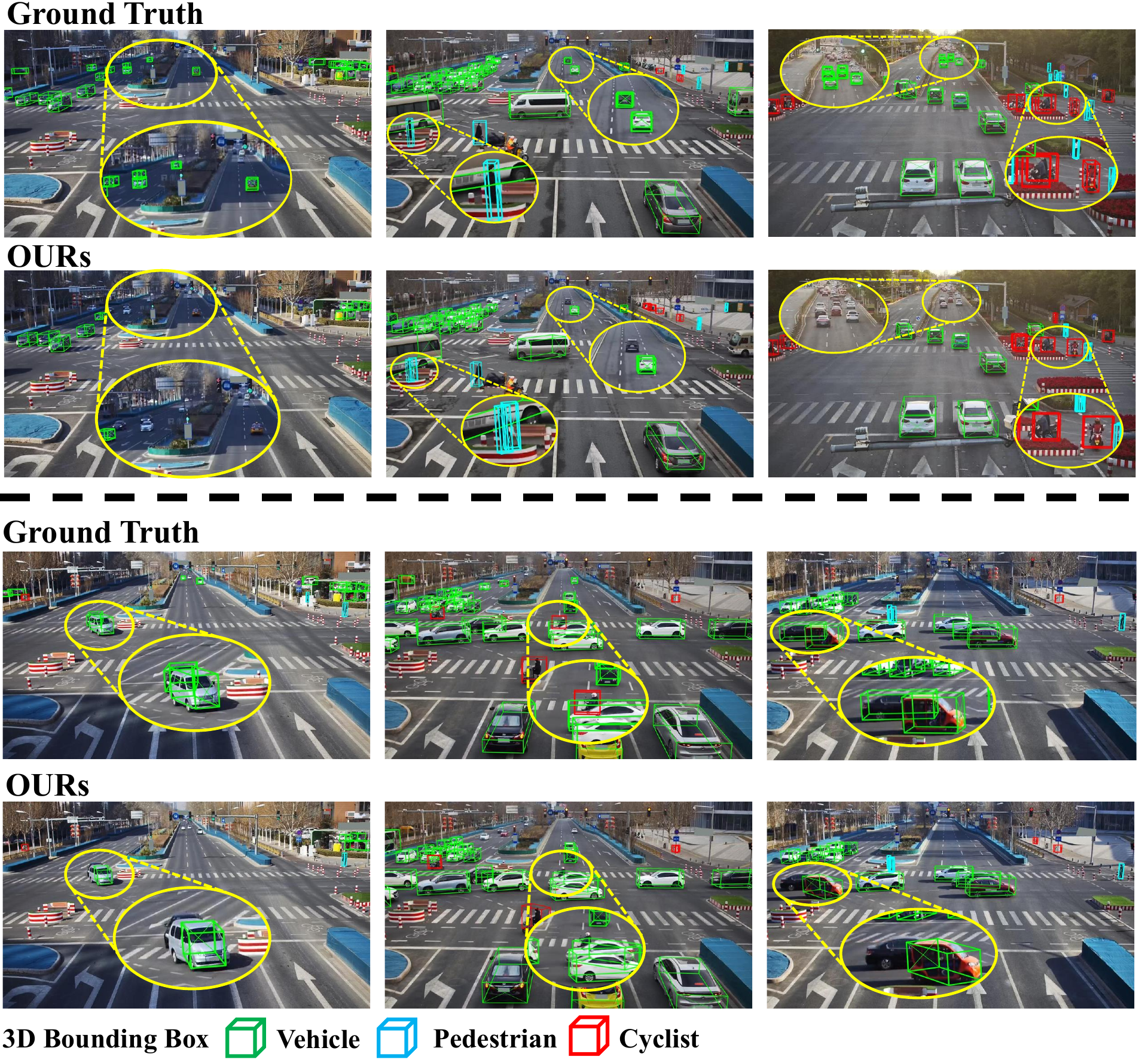}
\caption{\textbf{Failure cases of LH3D on the DAIR-V2X-I validation set: Ground-truth annotations vs. LH3D predictions.}
The top row mainly illustrates long-range perception, where distant vehicles provide limited visual cues, leading to missed detections or unstable 3D localization.
The bottom row shows failures caused by occlusion, where overlapping objects hinder geometric reasoning and result in incomplete predictions.
}
    \label{fig:failure-cases}
\end{figure*}

\subsection{Failure Cases}
\label{sec:failure}
Despite LH2D's strong performance over baseline methods, our approach still encounters failure cases in challenging roadside scenarios, particularly for distant vehicles and for pedestrians or cyclists that are occluded. 

Fig.~\ref{fig:failure-cases} highlights two primary failure modes: \textit{\textbf{distance}} and \textit{\textbf{occlusion}}.
First, long-range objects often lack sufficient visual detail for reliable 3D estimation, leading to missed detections of small or distant vehicles (\textit{top examples}). Cyclists also pose a challenge, as they are easily misclassified in crowded environments.

Second, the model struggles with severe occlusion (\textit{bottom examples}). When vehicles heavily overlap, LH3D frequently fails to distinguish the object in the rear. This issue extends to vulnerable road users; for instance, the visualization shows a cyclist largely screened by a vehicle, resulting in a missed detection due to the lack of visible features.

\subsection{Validation of Hierarchical Stages}
\label{sec:stage_validity}
\subsubsection{Stage 1: Depth-Confident Sample Selection}
\label{sec:stage1}
Stage~1 aims to filter out inherently ambiguous scenes by selecting images where the depth estimator exhibits high confidence and balanced depth coverage.
At the beginning of active learning, however, the detector is trained on only a very small labeled subset, so its depth predictions remain reliable only on relatively simple, low-ambiguity scenes.
As a result, Stage~1 naturally gravitates toward such scenes in the early rounds.
These early-selected images typically contain fewer objects, involve minimal occlusion, and show a more even spread of near- and mid-range targets.
In practical terms, the selected scenes also tend to have a lower density of vehicles, pedestrians, and cyclists, which prevents the annotation process from spending its limited early budget on congested or difficult scenes that the model is not yet strong enough to learn from.

As the active learning process progresses, the detector becomes increasingly capable of producing confident depth estimates on more complex layouts.
Stage~1 correspondingly begins to admit scenes with richer object arrangements, heavier occlusion, and greater geometric variability.
At the same time, the balanced-depth-coverage criterion avoids repeatedly sampling near-range scenes: once these bins are sufficiently covered, the objective encourages selecting images that contribute to underrepresented mid- and far-range regions.

Empirically, this behavior is clearly reflected on the DAIR-V2X-I~\cite{Yu2022DAIRV2X} dataset: LH3D consistently selects scenes with systematically closer and more learnable object configurations.
The average distance from annotated objects to the camera is \textbf{6.84\,m} under LH3D, whereas uncertainty-based and diversity-based baselines select scenes whose average distance consistently exceeds \textbf{7.5\,m}.
This demonstrates that LH3D not only favors depth-confident, easy-to-learn scenes in early rounds but also preserves annotation budget by postponing dense or ambiguous scenes until the model becomes sufficiently strong to extract reliable supervision from them.

\begin{figure*}[t!]
    \centering \includegraphics[width=\textwidth]{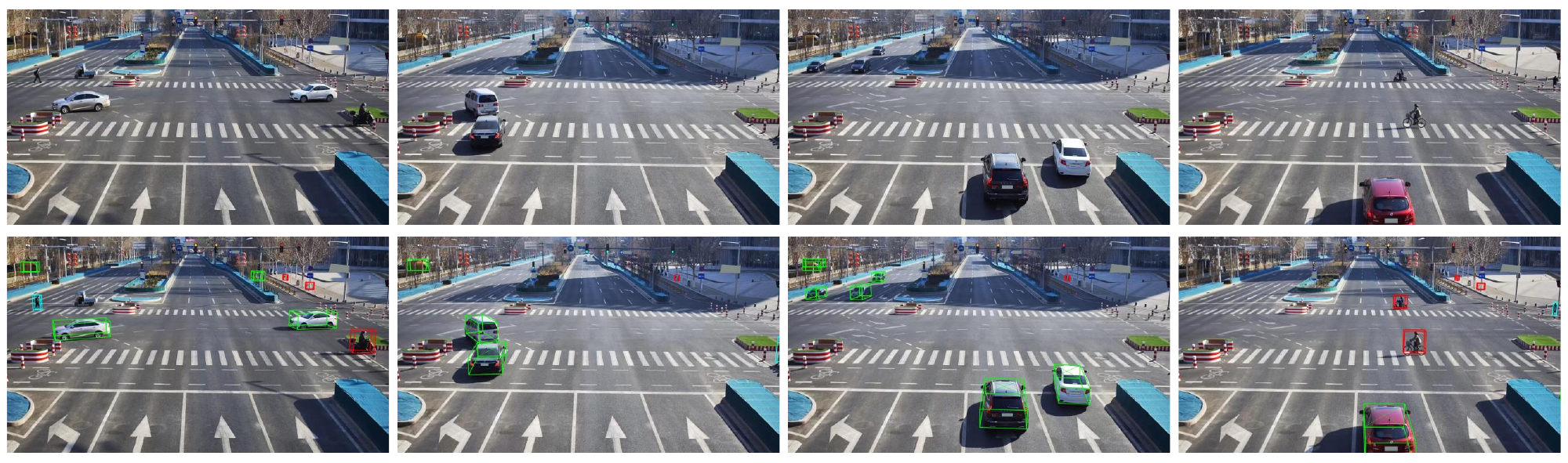}
\caption{\textbf{Training samples from DAIR-V2X-I selected during LH3D Stage 1 through depth-confident sample selection.} The top row displays the original images, and the bottom row shows the corresponding 3D bounding box annotations. The samples selected during Stage 1 are characterized by high visual clarity and minimal ambiguity. The selection strategy prioritizes scenes where vehicles, pedestrians, and cyclists appear without occlusion and are positioned at moderate distances. Furthermore, these samples exhibit low scene complexity, avoiding overcrowded traffic environments.
}
    \label{fig:selected}
\end{figure*}

\subsubsection{Stage 2: Rare-Common Class Balancing}
\label{sec:stage2}

The core function of Stage 2 is to ensure that the selected annotation set maintains high class diversity across multiple active learning rounds, especially under the severe imbalance in roadside 3D datasets (e.g., vehicles vastly outnumber pedestrians and cyclists).

\begin{figure}[t!]
    \centering
    \includegraphics[width=0.8\columnwidth]{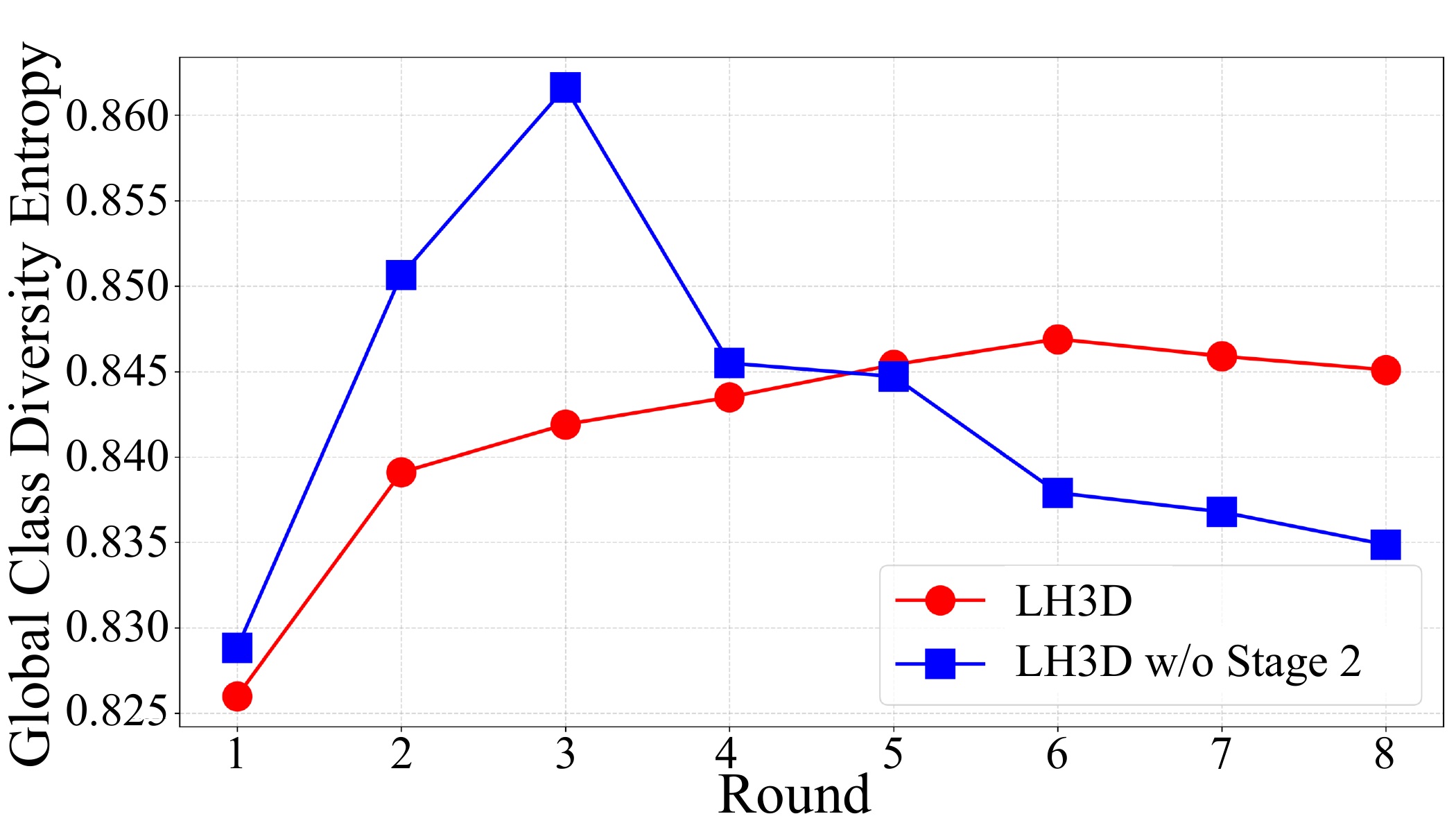}
\caption{
Global class-diversity entropy over AL rounds: comparison between LH3D and LH3D without Stage 2.
}
\label{fig:stage2-entropy}
\end{figure}

To validate the necessity of Stage 2, we perform an ablation study comparing the full LH3D pipeline against LH3D w/o Stage 2.
As shown in Fig.~\ref{fig:stage2-entropy}, we track the global class-diversity entropy over 8 active learning rounds (higher entropy indicates better class balance).

Initially, removing Stage 2 (blue curve) results in slightly higher entropy than the full LH3D pipeline (red curve), peaking around Round 3 (0.862).
This surge occurs because, without explicit balancing constraints, the baseline aggressively selects available rare classes (Pedestrians and Cyclists) from the unlabeled pool, leading to a temporary increase in diversity.

However, this high diversity is unsustainable.
Since naturally rare classes are quickly depleted in early rounds, the LH3D w/o Stage 2 variant is forced to select mostly common classes (Vehicles) in later rounds, causing entropy to drop significantly (down to $\approx 0.835$ by Round 8).

In contrast, the full LH3D pipeline (with Stage 2) enforces a controlled, stable selection across classes.
Although its diversity gain is more gradual at the beginning, it maintains high and stable class diversity throughout the process, stabilizing at an entropy of $\approx 0.845$ in later rounds.
This consistent balancing prevents the selected set from becoming overly biased toward Vehicles and ultimately leads to better final detection performance: incorporating Stage 2 improves $AP_{3D|R40}$ by more than 4 points for Cyclists and 2 points for Pedestrians compared to LH3D w/o Stage 2.

\subsubsection{Stage 3: Geometric Variant Selection}
\label{sec:suppl-stage-c}

Stage~3 is designed to enhance per-class geometric \emph{variation} while still respecting the detector’s learned geometric priors, i.e., to select scenes that are novel but not extreme outliers in BEV layout space.

To test this design choice, we construct an ablated variant that inverts the first step of Stage~3: instead of favoring scenes whose BEV layouts are moderately consistent with the labeled set, it explicitly prioritizes scenes whose geometry is as \emph{dissimilar} as possible from previously labeled scenes.
In other words, we remove the geometric consistency constraint and aggressively push selection toward maximal geometric novelty.

On DAIR-V2X-I (Hard setting) with a BEVHeight backbone and the same total annotation budget, this “maximally dissimilar” variant yields substantially worse performance: the final $AP_{3D|R40}$ averaged over Car, Pedestrian, and Cyclist is lower by about 4 percent compared to the full LH3D with Stage~3.
Qualitatively, the ablated variant tends to oversample rare, highly anomalous layouts in early rounds, which slows down training and introduces instability, as the model struggles to extract reliable supervision from overly difficult scenes.
In contrast, the original Stage~3, which encourages \emph{controlled} geometric deviation, maintains stable training dynamics and consistently achieves higher final detection accuracy.

\subsection{Ablation Studies on Annotation Budgets}
\label{sec:budget_ablation}

We conduct extensive ablation studies to evaluate the effectiveness of LH3D under varying annotation budgets on the DAIR-V2X-I dataset. The total available training pool contains $246,500$ objects (50\% of the total 493k annotations). The annotation budget is defined by the cumulative number of objects annotated. The budgets presented in Table~\ref{tab:dair-budgets} reflect the following proportions of the total training pool:
\begin{itemize}
    \item 8,000 objects: $\approx 3.24\%$ of the training pool.
    \item 16,000 objects: $\approx 6.49\%$ of the training pool.
    \item 24,000 objects: $\approx 9.74\%$ of the training pool.
    \item 32,000 objects: $\approx 12.98\%$ of the training pool.
\end{itemize}
Based on these findings, we chose the 32,000 object budget ($\approx 13\%$ of the training pool) as the primary comparative budget in the main text. At this level, our method, LH3D, achieves $86.06\%$, $67.32\%$, and $78.67\%$ of full-performance for vehicles, pedestrians, and cyclists respectively, significantly outperforming baselines and confirming that learnability, not uncertainty, matters for roadside 3D perception.

\begin{table*}[t!]
\centering
\tiny
\vspace{-5pt}
\caption{\textbf{\boldmath{$AP_{3D|R40}$} performance on the DAIR-V2X-I validation set} under different annotation budgets. The backbone detector is \textit{BEVHeight}.}
\label{tab:dair-budgets}
\setlength{\tabcolsep}{4.5pt}
\begin{threeparttable}
\resizebox{\textwidth}{!}{
\renewcommand{\arraystretch}{0.9}
\begin{tabular}{l ccc ccc ccc ccc}
\toprule
& \multicolumn{3}{c}{\textbf{Vehicle (\textit{IoU=0.5})}} &
  \multicolumn{3}{c}{\textbf{Pedestrian (\textit{IoU=0.25})}} &
  \multicolumn{3}{c}{\textbf{Cyclist (\textit{IoU=0.25})}} &
  \multicolumn{3}{c}{\textbf{Average}} \\
\cmidrule(lr){2-4}\cmidrule(lr){5-7}\cmidrule(lr){8-10}\cmidrule(lr){11-13}
\textbf{Method (Object Budget)} &
Easy & Mod. & Hard &
Easy & Mod. & Hard &
Easy & Mod. & Hard &
Easy & Mod. & Hard \\
\midrule
RANDOM (8000)    & 51.95 & 43.94 & 43.91 & 14.01 & 13.28 & 13.45 & 23.42 & 30.56 & 31.07 & 29.79 & 29.26 & 29.48 \\
ENTROPY (8000)   & 51.85 & 42.56 & 42.64 & \textbf{14.96} & \textbf{14.13} & \textbf{14.27} & 22.74 & 33.60 & 34.07 & 29.85 & 30.10 & 30.33 \\
PPAL (8000)      & 50.18 & 42.40 & 42.49 & 13.00 & 12.19 & 12.28 & \textbf{26.73} & \textbf{39.33} & \textbf{39.47} & \textbf{29.97} & \textbf{31.31} & \textbf{31.41} \\
\rowcolor{gray!10}
LH3D (8000)      & \textbf{57.47} & \textbf{49.85} & \textbf{49.93} & 10.90 & 10.72 & 10.83 & 21.10 & 32.43 & 32.41 & 29.82 & 31.00 & 31.06 \\
\midrule
RANDOM (16000)   & 56.46 & 46.63 & 46.61 & 12.06 & 11.30 & 11.38 & 23.03 & 31.40 & 31.89 & 30.52 & 29.78 & 29.96 \\
ENTROPY (16000)  & 59.24 & 50.59 & 50.67 & 14.41 & 13.63 & 13.80 & 24.82 & 32.30 & 32.72 & 32.82 & 32.17 & 32.40 \\
PPAL (16000)     & 60.07 & 51.23 & 51.28 & 13.61 & 12.81 & 12.90 & \textbf{28.99} & \textbf{36.61} & \textbf{37.03} & 34.22 & 33.55 & 33.74 \\
\rowcolor{gray!10}
LH3D (16000)     & \textbf{63.03} & \textbf{52.93} & \textbf{52.41} & \textbf{15.81} & \textbf{14.84} & \textbf{14.94} & 27.64 & 33.64 & 34.30 & \textbf{35.49} & \textbf{33.80} & \textbf{33.88} \\
\midrule
RANDOM (24000)   & 57.36 & 48.88 & 48.97 & 13.50 & 12.98 & 12.97 & 26.66 & \textbf{37.36} & \textbf{37.53} & 32.51 & 33.07 & 33.16 \\
ENTROPY (24000)  & 57.44 & 48.99 & 49.14 & 13.58 & 12.73 & 12.83 & 27.68 & 34.27 & 34.69 & 32.90 & 32.00 & 32.22 \\
PPAL (24000)     & 58.31 & 48.14 & 48.27 & 11.98 & 11.69 & 11.83 & 29.01 & 35.35 & 35.64 & 33.10 & 31.73 & 31.91 \\
\rowcolor{gray!10}
LH3D (24000)     & \textbf{63.55} & \textbf{53.71} & \textbf{52.78} & \textbf{16.58} & \textbf{15.65} & \textbf{15.82} & \textbf{29.93} & 36.95 & 37.33 & \textbf{36.69} & \textbf{35.44} & \textbf{35.31} \\
\midrule
RANDOM (32000)   & 61.90 & 51.37 & 51.41 & 13.63 & 13.23 & 13.42 & 30.04 & 38.70 & 39.38 & 35.19 & 34.43 & 34.74 \\
ENTROPY (32000)  & 63.42 & 54.42 & 54.51 & 17.50 & 16.57 & 16.72 & 31.45 & 36.86 & 38.57 & 37.46 & 36.67 & 36.53 \\
PPAL (32000)     & 60.20 & 51.38 & 51.44 & \textbf{19.09} & \textbf{18.47} & \textbf{18.07} & \textbf{34.41} & 39.13 & 39.71 & 37.90 & 36.33 & 36.41 \\
\rowcolor{gray!10}
LH3D (32000)     & \textbf{65.36} & \textbf{56.00} & \textbf{56.03} & 18.51 & 17.50 & 17.67 & 32.44 & \textbf{41.49} & \textbf{41.79} & \textbf{38.77} & \textbf{38.33} & \textbf{38.50} \\
\midrule
\rowcolor{blue!10}
ORACLE (246500)  & 73.05 & 61.32 & 61.19 & 22.10 & 21.57 & 21.11 & 42.85 & 42.26 & 42.09 & 46.00 & 41.72 & 41.46 \\
\bottomrule
\end{tabular}
}
\end{threeparttable}
\end{table*}

\subsection{Generalization Experiments}
\label{sec:rope3D}
Table~\ref{tab:rope3d_results} demonstrates the effectiveness of our proposed LH3D method on the Rope3D dataset. 
Across both \textit{BEVSpread} and \textit{BEVDet} backbones, LH3D consistently surpasses state-of-the-art active learning baselines. 
In particular, under the \textit{BEVSpread} configuration, our method outperforms the PPAL baseline by more than 2.6 points in the \textit{Easy Vehicle} category.
For the \textit{Cyclist} category on the \textit{Hard} difficulty setting (using the BEVSpread backbone), LH3D achieves an $AP_{3D}$ of 17.65. 
This represents a substantial improvement of +3.61 points over the nearest competitor, PPAL ($14.04$), highlighting our model's effectiveness in mitigating the ambiguities often associated with vulnerable road users.

We observe that the performance on the Rope3D dataset is lower compared to DAIR-V2X-I. This discrepancy can be attributed to the higher complexity of the Rope3D scenarios and the limited scale of the validation set ($1,688$ images), which poses a greater challenge for the model under the current active learning constraints. In future work, we will increase the annotation budget to select a larger number of informative samples for training, thereby further improving the model's generalization capability.

\begin{table*}[t!]
\vspace{-10pt}
\tiny
\centering
\caption{\textbf{\boldmath{$AP_{3D|R40}$} results on the Rope3D validation set} with 20\% queried boxes. Backbones include  \textit{BEVSpread} and \textit{BEVDet}.}
\label{tab:rope3d_results}
\setlength{\tabcolsep}{6pt}
\scriptsize
\begin{threeparttable}
\resizebox{\textwidth}{!}
{
\renewcommand{\arraystretch}{0.9}
\begin{tabular}{l l ccc ccc ccc ccc}
\toprule
& & \multicolumn{3}{c}{\textbf{Vehicle (IoU=0.5)}} &
    \multicolumn{3}{c}{\textbf{Pedestrian (IoU=0.25)}} &
    \multicolumn{3}{c}{\textbf{Cyclist (IoU=0.25)}} &
    \multicolumn{3}{c}{\textbf{Average}} \\
\cmidrule(lr){3-5}\cmidrule(lr){6-8}\cmidrule(lr){9-11}\cmidrule(lr){12-14}
\textbf{Backbone} & \textbf{Method} &
Easy & Mod. & Hard &
Easy & Mod. & Hard &
Easy & Mod. & Hard &
Easy & Mod. & Hard \\
\midrule
\multirow{8}{*}{\textbf{BEVSpread}}
& RANDOM       & 24.49 & 22.52 & 22.41 & 1.90 & 1.83 & 1.86 & 8.85 & 12.03 & 12.01 & 11.75 & 12.13 & 12.09\\
& ENTROPY      & 24.72 & 20.63 & 20.52 & 0.80 & 0.81 & 0.83 & 7.46 & 10.76 & 10.77 & 10.33 & 10.73 & 11.37 \\
& BGADL~\cite{tran2019bayesian}    & 24.13 & 23.89 & 22.13 & 1.13 & 1.05 & 1.39 & 13.23 & 12.00 & 12.11 & 12.83 & 12.31 & 11.88\\
& CORESET~\cite{sener2017active}  & 24.61 & 22.62 & 21.41 & 1.38 & 1.18 & 1.20 & 9.04 & 12.20 & 12.17 & 11.68 & 12.00 & 11.59\\
& BADGE~\cite{ash2019deep}         & 24.70 & 24.13 & 23.89 & 1.28 & 1.10 & 1.04 & 12.74 & 13.99 & 13.10 & 12.91 & 13.07 & 12.67\\
& PPAL~\cite{yang2024plug}        & 28.19 & 25.47 & 24.03 & 2.47 & 2.56 & 2.62 & 11.14 & 14.03 & 14.04 & 13.93 & 14.02 & 13.56\\
& HUA~\cite{park2023active}        & 23.99 & 22.17 & 20.87 & 1.96 & 1.86 & 1.89 & 7.91 & 11.65 & 11.63 & 11.29 & 11.89 & 11.46\\
\rowcolor{gray!10}
& \textbf{LH3D (Ours)}    & \textbf{30.85} & \textbf{26.75} & \textbf{26.60} & \textbf{2.53} & \textbf{2.53} & \textbf{2.57} & \textbf{14.30} & \textbf{17.69} & \textbf{17.65} & \textbf{15.89} & \textbf{15.66} & \textbf{15.61}\\
\midrule

\multirow{8}{*}{\textbf{BEVDet}}
& RANDOM       & 23.50 & 21.73 & 21.02 & 1.57 & 1.71 & 1.72 & 7.58 & 13.18 & 13.27 & 10.88 & 12.21 & 11.99\\
& ENTROPY      & 25.84 & 22.64 & 22.62 & 1.12 & 1.16 & 1.18 & 9.46 & 13.85 & 13.07 & 12.14 & 12.55 & 12.29\\
& BGADL~\cite{tran2019bayesian}    & 23.25 & 20.17 & 20.71 & 1.02 & 1.01 & 1.08 & 9.41 & 11.66 & 11.47 & 11.23 & 10.95 & 11.09\\
& CORESET~\cite{sener2017active}  & 26.26 & 22.61 & 22.59 & 1.80 & 1.68 & 1.72 & 10.63 & 14.13 & 14.21 & 12.90 & 12.81 & 12.84\\
& BADGE~\cite{ash2019deep}         & 24.77 & 22.70 & 21.03 & 1.81 & 1.61 & 1.90 & \textbf{11.72} & 12.63 & 12.28 & 12.77 & 12.31 & 11.74\\
& PPAL~\cite{yang2024plug}        & 21.53 & 20.29 & 18.96 & 1.78 & 1.78 & 1.80 & 7.20 & 10.43 & 10.43 & 10.17 & 10.83 & 10.40\\
& HUA~\cite{park2023active}        & 21.39 & 20.29 & 20.22 & 1.26 & 1.15 & 1.16 & 6.81 & 10.36 & 10.39 & 9.82 & 10.60 & 10.59\\
\rowcolor{gray!10}
& \textbf{LH3D (Ours)}& \textbf{28.19} & \textbf{26.09} & \textbf{25.97} & \textbf{1.78} & \textbf{1.84} & \textbf{1.90} & 11.59 & \textbf{16.37} & \textbf{16.44} & \textbf{13.85} & \textbf{14.76 }& \textbf{14.77}\\
\bottomrule
\end{tabular}
}
\end{threeparttable}

\end{table*}

\subsection{Computational Complexity}
\label{sec:complexity}
We evaluate the computational efficiency of our proposed approach by comparing the training duration against several baseline methods. Table~\ref{tab:complexity} presents the training time comparison on the DAIR-V2X-I dataset with the \textit{BEVHeight} backbone.

While the \textsc{Random} strategy achieves the lowest training time ($3.55$ hours) due to its lack of selection overhead, our method, LH3D, maintains competitive efficiency. With a total training time of $4.47$ hours, LH3D proves to be more efficient than both PPAL ($4.70$ hours). 

\begin{table}[h]
\centering
\small
\setlength{\tabcolsep}{6pt}
\caption{\textbf{Training Time Comparison on DAIR-V2X-I} using the \textit{BEVHeight} backbone.}
\label{tab:complexity}
\begin{threeparttable}
\begin{tabular}{lc}
\toprule
\textbf{Method} & \textbf{Time (\textit{hours})} \\
\midrule
RANDOM  &  3.55  \\
ENTROPY &  4.40 \\
PPAL &  4.70 \\
HUA &  4.08 \\
\rowcolor{gray!10}\textbf{LH3D (Ours)} & 4.47 \\
\bottomrule
\end{tabular}
\end{threeparttable}
\end{table}

\subsection{Extended Analysis: Human Study on Ambiguity}
\label{sec:human_study}

To empirically validate our hypothesis that \emph{inherently ambiguous samples} provide weaker supervision signals than learnable samples—even when accurate ground truth is provided—we conducted a controlled human study. This study isolates the impact of visual ambiguity from other factors like class imbalance or label noise.

\subsubsection{Study Setup and Partitioning}
We enlisted three expert annotators (well-trained PhD students in the computer vision domain) to manually partition the unlabeled training pool into two distinct subsets: \textbf{Learnable} and \textbf{Ambiguous}. The classification was based on three primary visual criteria strictly from a monocular perspective:
\begin{itemize}
    \item \textbf{Object Distance:} Scenes dominated by objects at extreme ranges (e.g., $>55$m) where objects have lower resolution than closer objects.
    \item \textbf{Occlusion Level:} Scenes where key objects suffer from severe occlusion ($>70\%$) or are truncated.
    \item \textbf{Scene Clutter:} High-density scenes where object boundaries are visually indistinguishable.
\end{itemize}
To ensure a fair comparison, the annotators strictly controlled the selection to maintain a consistent class distribution (Car, Pedestrian, Cyclist) between the two subsets, eliminating semantic imbalance as a confounding variable.

\subsubsection{Experimental Protocol}
We designed an iterative training protocol to mimic the active learning process, but with manual selection:
\begin{itemize}
    \item \textbf{Budget Constraints:} The total annotation budget was fixed at $10,000$ objects.
    \item \textbf{Iterative Selection:} The process spanned 10 rounds. In each round, annotators selected up to 50 images from their respective pools (Learnable vs. Ambiguous) to add to the training set.
    \item \textbf{Training Settings:} The model was trained for 10 epochs per round. To simulate a realistic active learning cycle, the model for round $k$ was initialized with the weights from round $k-1$ (incremental learning).
    \item \textbf{Labeling:} Both groups were trained using the official Ground Truth labels from the dataset.
\end{itemize}

\subsubsection{Results and Discussion}
The results, visualized in Fig.~\ref{fig:ambiguous_vs_easy} of the main paper, reveal a critical finding:

\textbf{Ambiguity Limits Monocular Learnability.} Despite using the exact same detector architecture, optimizer, and reliable ground truth labels, the model trained on the \emph{Ambiguous} split consistently underperformed the model trained on the \emph{Learnable} split.
Specifically, the performance gap is most pronounced for Vehicles and Pedestrians. This indicates that ambiguous samples suffer from low signal-to-noise ratios; even with correct labels, the image features (due to blur or occlusion) are insufficient for the network to learn a generalized geometric mapping.

\textbf{Implication for Active Learning.} This experiment confirms that in the roadside monocular setting, \emph{uncertainty} is not equivalent to \emph{informativeness}. High-uncertainty samples in this domain are often inherently ambiguous cases that confuse the model rather than strengthen it. This validates the core motivation of LH3D: prioritizing learnability over mere uncertainty.

%% file: main.bbl
\begin{thebibliography}{42}
\providecommand{\natexlab}[1]{#1}
\providecommand{\url}[1]{\texttt{#1}}
\expandafter\ifx\csname urlstyle\endcsname\relax
  \providecommand{\doi}[1]{doi: #1}\else
  \providecommand{\doi}{doi: \begingroup \urlstyle{rm}\Url}\fi

\bibitem[Agarwal et~al.(2020)Agarwal, Arora, Anand, and
  Arora]{agarwal2020contextual}
Sharat Agarwal, Himanshu Arora, Saket Anand, and Chetan Arora.
\newblock Contextual diversity for active learning.
\newblock In \emph{European Conference on Computer Vision}, pages 137--153.
  Springer, 2020.

\bibitem[Ash et~al.(2019)Ash, Zhang, Krishnamurthy, Langford, and
  Agarwal]{ash2019deep}
Jordan~T Ash, Chicheng Zhang, Akshay Krishnamurthy, John Langford, and Alekh
  Agarwal.
\newblock Deep batch active learning by diverse, uncertain gradient lower
  bounds.
\newblock \emph{arXiv preprint arXiv:1906.03671}, 2019.

\bibitem[Bilmes(2022)]{bilmes2022submodularity}
Jeff Bilmes.
\newblock Submodularity in machine learning and artificial intelligence.
\newblock \emph{arXiv preprint arXiv:2202.00132}, 2022.

\bibitem[Caesar et~al.(2019)Caesar, Bankiti, Lang, Vora, Liong, Xu, Krishnan,
  Pan, Baldan, and Beijbom]{nuscenes2019}
Holger Caesar, Varun Bankiti, Alex~H. Lang, Sourabh Vora, Venice~Erin Liong,
  Qiang Xu, Anush Krishnan, Yu Pan, Giancarlo Baldan, and Oscar Beijbom.
\newblock nuscenes: A multimodal dataset for autonomous driving.
\newblock \emph{arXiv preprint arXiv:1903.11027}, 2019.

\bibitem[Cohn et~al.(1996)Cohn, Ghahramani, and Jordan]{cohn1996active}
David~A Cohn, Zoubin Ghahramani, and Michael~I Jordan.
\newblock Active learning with statistical models.
\newblock \emph{Journal of artificial intelligence research}, 4:\penalty0
  129--145, 1996.

\bibitem[Gal and Ghahramani(2015)]{gal2015bayesian}
Yarin Gal and Zoubin Ghahramani.
\newblock Bayesian convolutional neural networks with bernoulli approximate
  variational inference.
\newblock \emph{arXiv preprint arXiv:1506.02158}, 2015.

\bibitem[Geiger et~al.(2012)Geiger, Lenz, and Urtasun]{Geiger2012CVPR}
Andreas Geiger, Philip Lenz, and Raquel Urtasun.
\newblock Are we ready for autonomous driving? the kitti vision benchmark
  suite.
\newblock In \emph{Proceedings of the IEEE Conference on Computer Vision and
  Pattern Recognition (CVPR)}, pages 3354--3361, 2012.

\bibitem[Houlsby et~al.(2011)Houlsby, Husz{\'a}r, Ghahramani, and
  Lengyel]{houlsby2011bayesian}
Neil Houlsby, Ferenc Husz{\'a}r, Zoubin Ghahramani, and M{\'a}t{\'e} Lengyel.
\newblock Bayesian active learning for classification and preference learning.
\newblock \emph{arXiv preprint arXiv:1112.5745}, 2011.

\bibitem[Huang et~al.(2021)Huang, Huang, Zhu, Ye, and Du]{huang2021bevdet}
Junjie Huang, Guan Huang, Zheng Zhu, Yun Ye, and Dalong Du.
\newblock Bevdet: High-performance multi-camera 3d object detection in
  bird-eye-view.
\newblock \emph{arXiv preprint arXiv:2112.11790}, 2021.

\bibitem[Kim et~al.(2021)Kim, Park, Kim, and Chun]{kim2021task}
Kwanyoung Kim, Dongwon Park, Kwang~In Kim, and Se~Young Chun.
\newblock Task-aware variational adversarial active learning.
\newblock In \emph{Proceedings of the IEEE/CVF conference on computer vision
  and pattern recognition}, pages 8166--8175, 2021.

\bibitem[Krause et~al.(2008)Krause, Singh, and Guestrin]{krause2008near}
Andreas Krause, Ajit Singh, and Carlos Guestrin.
\newblock Near-optimal sensor placements in gaussian processes: Theory,
  efficient algorithms and empirical studies.
\newblock \emph{Journal of Machine Learning Research}, 9\penalty0 (8):\penalty0
  235--284, 2008.

\bibitem[Lewis(1995)]{lewis1995sequential}
David~D Lewis.
\newblock A sequential algorithm for training text classifiers: Corrigendum and
  additional data.
\newblock In \emph{Acm Sigir Forum}, pages 13--19. ACM New York, NY, USA, 1995.

\bibitem[Li et~al.(2022{\natexlab{a}})Li, Ge, Yu, Yang, Wang, Shi, Sun, and
  Li]{li2022bevdepth}
Yinhao Li, Zheng Ge, Guanyi Yu, Jinrong Yang, Zengran Wang, Yukang Shi,
  Jianjian Sun, and Zeming Li.
\newblock Bevdepth: Acquisition of reliable depth for multi-view 3d object
  detection.
\newblock \emph{arXiv preprint arXiv:2206.10092}, 2022{\natexlab{a}}.

\bibitem[Li et~al.(2022{\natexlab{b}})Li, Ma, An, Wang, Zhong, Chen, and
  Feng]{li2022v2x}
Yiming Li, Dekun Ma, Ziyan An, Zixun Wang, Yiqi Zhong, Siheng Chen, and Chen
  Feng.
\newblock V2x-sim: Multi-agent collaborative perception dataset and benchmark
  for autonomous driving.
\newblock \emph{IEEE Robotics and Automation Letters}, 7\penalty0 (4):\penalty0
  10914--10921, 2022{\natexlab{b}}.

\bibitem[Li et~al.(2022{\natexlab{c}})Li, Wang, Li, Xie, Sima, Lu, Qiao, and
  Dai]{li2022bevformer}
Zhiqi Li, Wenhai Wang, Hongyang Li, Enze Xie, Chonghao Sima, Tong Lu, Yu Qiao,
  and Jifeng Dai.
\newblock Bevformer: Learning bird's-eye-view representation from multi-camera
  images via spatiotemporal transformers.
\newblock \emph{arXiv preprint arXiv:2203.17270}, 2022{\natexlab{c}}.

\bibitem[Lin(1991)]{lin1991divergence}
Jianhua Lin.
\newblock Divergence measures based on the shannon entropy.
\newblock \emph{IEEE Transactions on Information theory}, 37\penalty0
  (1):\penalty0 145--151, 1991.

\bibitem[Mac~Aodha et~al.(2014)Mac~Aodha, Campbell, Kautz, and
  Brostow]{mac2014hierarchical}
Oisin Mac~Aodha, Neill~DF Campbell, Jan Kautz, and Gabriel~J Brostow.
\newblock Hierarchical subquery evaluation for active learning on a graph.
\newblock In \emph{Proceedings of the IEEE conference on computer vision and
  pattern recognition}, pages 564--571, 2014.

\bibitem[Mirzasoleiman et~al.(2015)Mirzasoleiman, Badanidiyuru, Karbasi,
  Vondr{\'a}k, and Krause]{mirzasoleiman2015lazier}
Baharan Mirzasoleiman, Ashwinkumar Badanidiyuru, Amin Karbasi, Jan Vondr{\'a}k,
  and Andreas Krause.
\newblock Lazier than lazy greedy.
\newblock In \emph{Proceedings of the AAAI Conference on Artificial
  Intelligence (AAAI)}, pages 1812--1818, 2015.

\bibitem[Nemhauser et~al.(1978)Nemhauser, Wolsey, and
  Fisher]{nemhauser1978analysis}
George~L. Nemhauser, Laurence~A. Wolsey, and Marshall~L. Fisher.
\newblock An analysis of approximations for maximizing submodular set
  functions---{I}.
\newblock \emph{Mathematical Programming}, 14\penalty0 (1):\penalty0 265--294,
  1978.

\bibitem[Park et~al.(2023)Park, Choi, Kim, Han, and Moon]{park2023active}
Younghyun Park, Wonjeong Choi, Soyeong Kim, Dong-Jun Han, and Jaekyun Moon.
\newblock Active learning for object detection with evidential deep learning
  and hierarchical uncertainty aggregation.
\newblock In \emph{The Eleventh International Conference on Learning
  Representations}, 2023.

\bibitem[Philion and Fidler(2020)]{philion2020lss}
Jonah Philion and Sanja Fidler.
\newblock Lift, splat, shoot: Encoding images from arbitrary camera rigs by
  implicitly unprojecting to 3d.
\newblock In \emph{Proceedings of the European Conference on Computer Vision
  (ECCV)}, 2020.

\bibitem[Roy et~al.(2018)Roy, Unmesh, and Namboodiri]{roy2018deep}
Soumya Roy, Asim Unmesh, and Vinay~P Namboodiri.
\newblock Deep active learning for object detection.
\newblock In \emph{BMVC}, page~91, 2018.

\bibitem[Sener and Savarese(2017)]{sener2017active}
Ozan Sener and Silvio Savarese.
\newblock Active learning for convolutional neural networks: A core-set
  approach.
\newblock \emph{arXiv preprint arXiv:1708.00489}, 2017.

\bibitem[Simonelli et~al.(2019)Simonelli, {Rota Bul{\`o}}, Porzi,
  L{\'o}pez-Antequera, and Kontschieder]{simonelli2019disentangling}
Andrea Simonelli, Samuel {Rota Bul{\`o}}, Lorenzo Porzi, Manuel
  L{\'o}pez-Antequera, and Peter Kontschieder.
\newblock Disentangling monocular {3D} object detection.
\newblock In \emph{Proceedings of the IEEE/CVF International Conference on
  Computer Vision}, pages 1991--1999, 2019.

\bibitem[Sinha et~al.(2019)Sinha, Ebrahimi, and Darrell]{sinha2019variational}
Samarth Sinha, Sayna Ebrahimi, and Trevor Darrell.
\newblock Variational adversarial active learning.
\newblock In \emph{Proceedings of the IEEE/CVF international conference on
  computer vision}, pages 5972--5981, 2019.

\bibitem[Sun et~al.(2020)Sun, Kretzschmar, Dotiwalla, Chouard, Patnaik, Tsui,
  Guo, Zhou, Chai, Caine, Vasudevan, Han, Ngiam, Zhao, Timofeev, Ettinger,
  Krivokon, Gao, Joshi, Zhao, Cheng, Zhang, Shlens, Chen, and
  Anguelov]{Sun_2020_CVPR}
Pei Sun, Henrik Kretzschmar, Xerxes Dotiwalla, Aurelien Chouard, Vijaysai
  Patnaik, Paul Tsui, James Guo, Yin Zhou, Yuning Chai, Benjamin Caine, Vijay
  Vasudevan, Wei Han, Jiquan Ngiam, Hang Zhao, Aleksei Timofeev, Scott
  Ettinger, Maxim Krivokon, Amy Gao, Aditya Joshi, Sheng Zhao, Shuyang Cheng,
  Yu Zhang, Jonathon Shlens, Zhifeng Chen, and Dragomir Anguelov.
\newblock Scalability in perception for autonomous driving: Waymo open dataset.
\newblock In \emph{Proceedings of the IEEE/CVF Conference on Computer Vision
  and Pattern Recognition (CVPR)}, 2020.

\bibitem[Tharwat and Schenck(2023)]{tharwat2023survey}
Alaa Tharwat and Wolfram Schenck.
\newblock A survey on active learning: State-of-the-art, practical challenges
  and research directions.
\newblock \emph{Mathematics}, 11\penalty0 (4):\penalty0 820, 2023.

\bibitem[Tran et~al.(2019)Tran, Do, Reid, and Carneiro]{tran2019bayesian}
Toan Tran, Thanh-Toan Do, Ian Reid, and Gustavo Carneiro.
\newblock Bayesian generative active deep learning.
\newblock In \emph{International conference on machine learning}, pages
  6295--6304. PMLR, 2019.

\bibitem[Wang and Shang(2014)]{wang2014new}
Dan Wang and Yi Shang.
\newblock A new active labeling method for deep learning.
\newblock In \emph{2014 International joint conference on neural networks
  (IJCNN)}, pages 112--119. IEEE, 2014.

\bibitem[Wang et~al.(2017)Wang, Zhang, Li, Zhang, and Lin]{wang2016cost}
Keze Wang, Dongyu Zhang, Ya Li, Ruimao Zhang, and Liang Lin.
\newblock Cost-effective active learning for deep image classification.
\newblock \emph{IEEE Transactions on Circuits and Systems for Video
  Technology}, 27\penalty0 (12):\penalty0 2591--2600, 2017.

\bibitem[Wang et~al.(2024)Wang, Lu, Zheng, Zhan, Ye, Tan, Wang, Wang, and
  Li]{Wang2024BEVSpread}
Wenjie Wang, Yehao Lu, Guangcong Zheng, Shuigen Zhan, Xiaoqing Ye, Zichang Tan,
  Jingdong Wang, Gaoang Wang, and Xi Li.
\newblock Bevspread: Spread voxel pooling for bird's-eye-view representation in
  vision-based roadside 3d object detection.
\newblock In \emph{Proceedings of the IEEE/CVF Conference on Computer Vision
  and Pattern Recognition (CVPR)}, pages 14718--14727, 2024.

\bibitem[Wei et~al.(2014)Wei, Liu, Kirchhoff, Bartels, and
  Bilmes]{wei2014submodular}
Kai Wei, Yuzong Liu, Katrin Kirchhoff, Chris Bartels, and Jeff Bilmes.
\newblock Submodular subset selection for large-scale speech training data.
\newblock In \emph{Proceedings of the IEEE International Conference on
  Acoustics, Speech and Signal Processing (ICASSP)}, pages 3311--3315, 2014.

\bibitem[Xu et~al.(2018)Xu, Ruiz{-}Juri, Huang, Duthie, and
  Clary]{Xu2018TrafficMonitoring}
Weijia Xu, Natalia Ruiz{-}Juri, Ruizhu Huang, Jennifer Duthie, and John Clary.
\newblock Automated pedestrian safety analysis using data from traffic
  monitoring cameras.
\newblock In \emph{Proceedings of the 1st ACM/EIGSCC Symposium on Smart Cities
  and Communities (SCC'18)}, pages 1--8. ACM, 2018.

\bibitem[Yang et~al.(2024)Yang, Huang, and Crowley]{yang2024plug}
Chenhongyi Yang, Lichao Huang, and Elliot~J Crowley.
\newblock Plug and play active learning for object detection.
\newblock In \emph{Proceedings of the IEEE/CVF conference on computer vision
  and pattern recognition}, pages 17784--17793, 2024.

\bibitem[Yang et~al.(2023)Yang, Yu, Tang, Li, Yuan, Wang, Zhang, and
  Chen]{Yang2023BEVHeight}
Lei Yang, Kaicheng Yu, Tao Tang, Jun Li, Kun Yuan, Li Wang, Xinyu Zhang, and
  Peng Chen.
\newblock Bevheight: A robust framework for vision-based roadside 3d object
  detection.
\newblock In \emph{Proceedings of the IEEE/CVF Conference on Computer Vision
  and Pattern Recognition (CVPR)}, 2023.

\bibitem[Yang et~al.(2015)Yang, Ma, Nie, Chang, and Hauptmann]{yang2015multi}
Yi Yang, Zhigang Ma, Feiping Nie, Xiaojun Chang, and Alexander~G Hauptmann.
\newblock Multi-class active learning by uncertainty sampling with diversity
  maximization.
\newblock \emph{International Journal of Computer Vision}, 113\penalty0
  (2):\penalty0 113--127, 2015.

\bibitem[Ye et~al.(2022)Ye, Shu, Li, Shi, Li, Wang, Tan, and
  Ding]{Ye2022Rope3D}
Xiaoqing Ye, Mao Shu, Hanyu Li, Yifeng Shi, Yingying Li, Guangjie Wang, Xiao
  Tan, and Errui Ding.
\newblock Rope3d: The roadside perception dataset for autonomous driving and
  monocular 3d object detection task.
\newblock In \emph{Proceedings of the IEEE/CVF Conference on Computer Vision
  and Pattern Recognition (CVPR)}, pages 21341--21350, 2022.

\bibitem[Yoo and Kweon(2019)]{yoo2019learning}
Donggeun Yoo and In~So Kweon.
\newblock Learning loss for active learning.
\newblock In \emph{Proceedings of the IEEE/CVF conference on computer vision
  and pattern recognition}, pages 93--102, 2019.

\bibitem[Yu et~al.(2022{\natexlab{a}})Yu, Luo, Shu, Huo, Yang, Shi, Guo, Li,
  Hu, Yuan, and Nie]{Yu2022DAIRV2X}
Haibao Yu, Yizhen Luo, Mao Shu, Yiyi Huo, Zebang Yang, Yifeng Shi, Zhenglong
  Guo, Hanyu Li, Xing Hu, Jirui Yuan, and Zaiqing Nie.
\newblock Dair-v2x: A large-scale dataset for vehicle-infrastructure
  cooperative 3d object detection.
\newblock In \emph{Proceedings of the IEEE/CVF Conference on Computer Vision
  and Pattern Recognition (CVPR)}, pages 21361--21370, 2022{\natexlab{a}}.

\bibitem[Yu et~al.(2022{\natexlab{b}})Yu, Zhu, Yang, and
  Chen]{yu2022consistency}
Weiping Yu, Sijie Zhu, Taojiannan Yang, and Chen Chen.
\newblock Consistency-based active learning for object detection.
\newblock In \emph{Proceedings of the IEEE/CVF conference on computer vision
  and pattern recognition workshops}, pages 3951--3960, 2022{\natexlab{b}}.

\bibitem[Zhu et~al.(2024)Zhu, Sheng, Cai, Deng, Yang, Liang, Chen, Gao, Song,
  and Ye]{Zhu2024RoScenes}
Xiaosu Zhu, Hualian Sheng, Sijia Cai, Bing Deng, Shaopeng Yang, Qiao Liang, Ken
  Chen, Lianli Gao, Jingkuan Song, and Jieping Ye.
\newblock Roscenes: A large-scale multi-view 3d dataset for roadside
  perception.
\newblock In \emph{Computer Vision -- ECCV 2024}, pages 331--347, 2024.

\bibitem[Zimmer et~al.(2023)Zimmer, Cre{\ss}, Nguyen, and
  Knoll]{zimmer2023tumtrafintersection}
Walter Zimmer, Christian Cre{\ss}, Huu~Tung Nguyen, and Alois~C. Knoll.
\newblock Tumtraf intersection dataset: All you need for urban 3d camera-lidar
  roadside perception.
\newblock In \emph{Proceedings of the 2023 IEEE 26th International Conference
  on Intelligent Transportation Systems (ITSC)}, pages 1030--1037. IEEE, 2023.

\end{thebibliography}
